%% file: arxiv.tex
\documentclass{article} 
\usepackage{arxiv,times}

\input{math_commands.tex}

\usepackage{hyperref}
\usepackage{url}

\usepackage{graphicx}
\usepackage{subcaption}
\usepackage{booktabs}
\usepackage{amssymb}
\usepackage{wrapfig}
\usepackage{multirow}
\usepackage{multicol}
\usepackage{makecell}
\usepackage{float}

\newcommand{\dqpar}[1]{\vspace{0.1em}\noindent\textbf{#1}}

\title{Disaggregated Quantization:\\ Specializing LLM Prefill and Decode}

\author{
Andrei Panferov\thanks{Work done during an internship at NVIDIA.}\\
NVIDIA \& ISTA
\And
Maximilian Kleinegger\\
ISTA
\And
Sweta Priyadarshi\\
NVIDIA
\AND
\begin{minipage}{\textwidth}
\centering\normalfont\normalsize
\begin{tabular}[t]{l}\textbf{Tijmen Blankevoort}\\ NVIDIA\end{tabular}
\hspace{4em}
\begin{tabular}[t]{l}\textbf{Dan Alistarh}\thanks{Correspondence to: \texttt{dan.alistarh@ist.ac.at}.}\\ ISTA\end{tabular}
\end{minipage}
}

\iclrfinalcopy 
\begin{document}

\maketitle

\begin{abstract}
Prefill and decode reward different approaches to quantization: low-precision arithmetic accelerates prompt processing, while compact weights reduce memory traffic during generation.
We propose ``disaggregated quantization'' (DQ), which specializes computation formats, weights and storage placement to both of these phases.
On Qwen 3 and Gemma 3, removing activation quantization specifically on decode improves accuracy on decode-heavy tasks without increasing inference cost. Training separate compute-native prefill weights accelerates prompt processing relative to weight-only inference while matching or exceeding its accuracy at 2--3-bit decode on both decode-heavy and prefill-heavy tasks.
With released Qwen3.8-27B GGUF decoders, training an NVFP4 prefiller improves 1-bit accuracy by $32.5$ points on MMLU-Pro and $35.3$ on MMMU-Pro without modifying the decode checkpoint.
To accommodate the additional checkpoint on a single device, offloaded disaggregated prefill (ODP) streams its weights from SSD, amortizing loading over prompt length.
On the same 27B model, ODP delivers a $1.78\times$ time-to-first-token speedup over the weight-only baseline at 8K prompt length in \texttt{llama.cpp}.
We evaluate accuracy under disaggregated serving in vLLM and further validate shared-weight format disaggregation through post-training quantization on models up to 2.8T parameters.
\end{abstract}

\section{Introduction}

Historically, transformer models were first proposed for sequence-to-sequence tasks~\citep{vaswani2023attentionneed}. In encoder--decoder transformers, the architecture for encoding and decoding is different. The encoder processes the input, while the decoder generates an output conditioned on its representations~\citep{raffel2023exploringlimitstransferlearning}. Cross-attention acts as a bridge between the two, allowing the two sub-networks to be trained toward the same output objective.

\begin{figure}[H]
    \centering
    \includegraphics[width=1.0\linewidth]{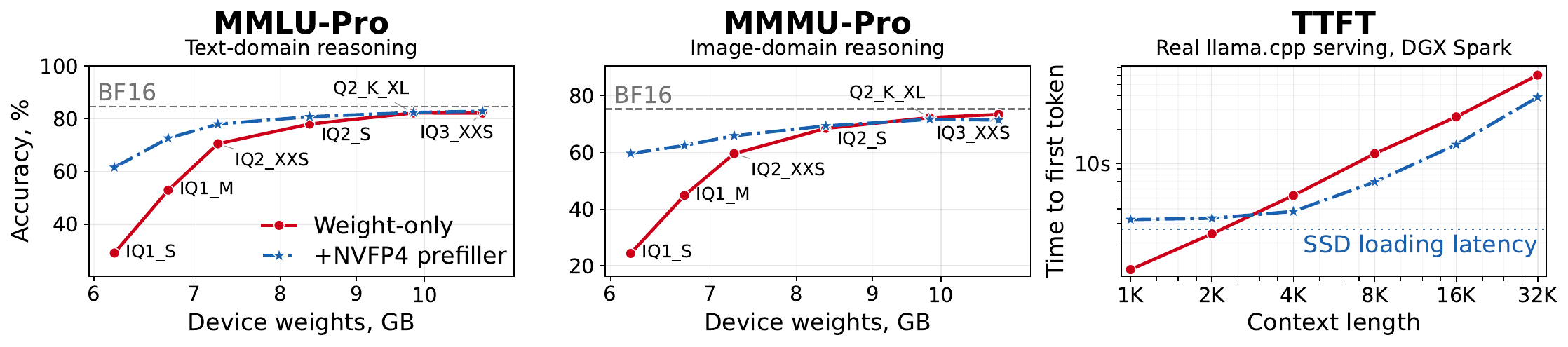}
    \caption{NVFP4 prefillers for off-the-shelf Qwen3.8-27B GGUF decoders. Left and middle: training the prefiller more than doubles IQ1\_S accuracy on both benchmarks while leaving the decode checkpoint unchanged. ODP adds no weight-memory overhead. Right: measured time to first token in \texttt{llama.cpp}, comparing ODP with weight-only IQ1\_S. ODP is faster from 4K context, reaching $1.78\times$ speedup at 8K.}
    \label{fig:pareto_gsq_rco_both}
\end{figure}

Modern decoder-only language models~\citep{Radford2019LanguageMA}, however, treat every token in a sequence equally as both a target conditioned on preceding tokens and context for future tokens.  Input processing and output generation consequently share the same parameters and architecture.

In instruction-following use-cases, the logical distinction reappears. A user supplies data or instructions, and the model produces an answer conditioned on them. Post-training reinforces these roles through structured interactions~\citep{wei2022finetunedlanguagemodelszeroshot,rafailov2024directpreferenceoptimizationlanguage} or rewards for generated answers~\citep{Guo_2025}. This splits up inference into two different phases where prefill constructs the prompt key--value (KV) cache and decode consumes it while generating the response. 

Prefill and decode workloads are different enough that large-scale deployment systems run them on separate accelerators. Similarly, we show that input processing and generation can use distinct quantized representations while remaining jointly optimized for the response. We explore this through \emph{disaggregated quantization} (DQ). DQ retains the pretrained attention architecture while specializing the linear computations and, optionally, their weights and their placement, to the two phases of LLM inference.

\subsection{Hardware cost}

\begin{figure*}[t]
    \centering
    \begin{subfigure}[t]{0.32\textwidth}
        \centering
        \includegraphics[width=\linewidth]{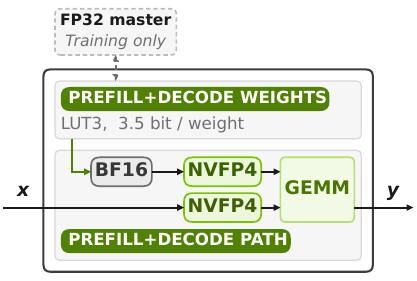}
        \caption{\textbf{Non-disaggregated} quantized linear. Common model weights, same formats for the two phases.}
        \label{fig:quantized_linear}
    \end{subfigure}
    \hfill
    \begin{subfigure}[t]{0.32\textwidth}
        \centering
        \includegraphics[width=\linewidth]{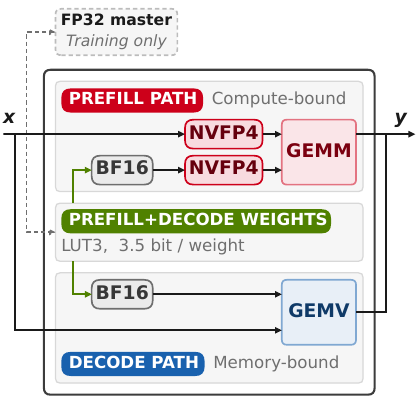}
        \caption{\textbf{Format-disaggregated} quantized linear. Common model weights, phase-specific formats.}
        \label{fig:dq_linear}
    \end{subfigure}
    \hfill
    \begin{subfigure}[t]{0.32\textwidth}
        \centering
        \includegraphics[width=\linewidth]{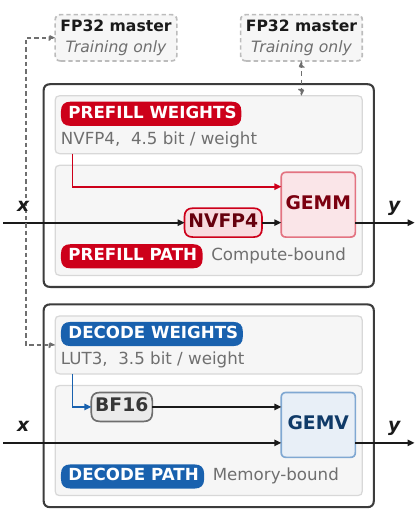}
        \caption{\textbf{Fully-disaggregated} quantized linear. Phase-specific model weights and formats.}
        \label{fig:full_disag_linear}
    \end{subfigure}
    \caption{Storage and computation schemes for various degrees of disaggregated quantization, using the LUT3 format as an example. This application of disaggregated quantization retains the prefill speed of NVFP4 and the decode speed of LUT3 while achieving higher accuracy (see Figure~\ref{fig:bars_disag_both}). Offloaded disaggregated prefill replaces full device residency of the extra prefill checkpoint with block buffers (see Section~\ref{sec:offloading}).}
    \label{fig:nvfp4_ablations}
\end{figure*}

\dqpar{Compute-bound prefill vs memory-bound decode.}
At every linear layer, inference combines (1) loading model weights and activations from device memory (DRAM) into the compute units and (2) general matrix multiplication (GEMM) inside them. For an $H\times H$ weight and $T\times H$ activations, it transfers $\mathcal{O}(H^2+TH)$ elements and performs $\mathcal{O}(TH^2)$ arithmetic. Increasing the token count $T$ therefore amortizes weight transfer over more computation. At batch one ($T=1$), each weight contributes only one multiply-add, so weight loading dominates. In sufficiently long prefill workloads ($T\gg1$), reuse across tokens makes the linear layers generally compute-bound.

\dqpar{Quantization formats target one of the two.}
These load profiles motivate (1) compressing weights to reduce memory traffic, admitting complex weight-only encodings~\citep{frantar2023gptqaccurateposttrainingquantization,egiazarian2024extremecompressionlargelanguage,tseng2024quipbetterllmquantization,tseng2025qtipquantizationtrellisesincoherence}, and (2) quantizing weights and activations to hardware-supported compute formats such as INT4~\citep{ashkboos2023quikendtoend4bitinference,ashkboos2024quarotoutlierfree4bitinference,liu2025spinquantllmquantizationlearned} or NVFP4~\citep{egiazarian2026bridginggappromiseperformance,chen2025intvsfpcomprehensive}.

A compact weight encoding, however, need not be a hardware-native compute format. A weight-only decode kernel can reconstruct values as it loads them for a matrix-vector product, still reducing weight traffic. It need not quantize activations either. Using arbitrary encoding with hardware-native quantized GEMM, however, requires both weight re-quantization and activation quantization into supported formats.

A common setup in efficient LLM inference is \textit{disaggregated serving}. Separate prefill and decode instances hold their own model weights and exchange a prompt KV cache, so decode must interpret representations produced by prefill. Keeping weights at each instance avoids transferring them between devices for every request; the interface between phases is instead the cache. Existing systems mostly address its communication~\citep{qin2025mooncakekvcachecentricdisaggregatedarchitecture} and scheduling~\citep{hu2024inferenceinterferencedisaggregatellm}.

\subsection{Contributions} 

\begin{figure}[t]
    \centering
    \includegraphics[width=1.0\linewidth]{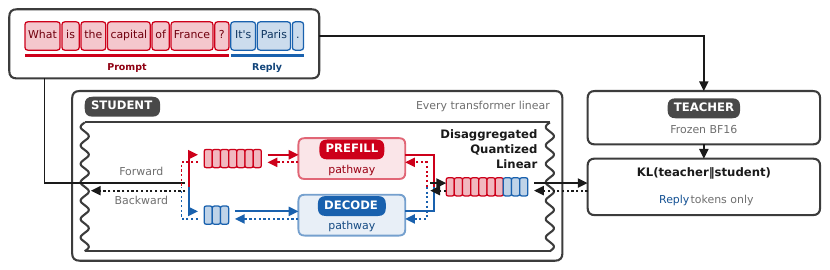}
    \caption{Quantization-aware distillation with disaggregation (QADD). The SFT assistant token mask, normally used only for loss, is also used to select the linear layer computational pathway.}
    \label{fig:qadd_training}
\end{figure}

We introduce \textit{disaggregated quantization} (DQ), a broad concept in which we treat quantization for prefill and decode separately. Splitting prefill and decode computation and weight formats yields a ladder of different schemes, each targeting a different axis of inference cost. We introduce the different schemes, and the tools to optimize networks for each. Our contributions are as follows:

\begin{enumerate}
    \item \textbf{Quantization-aware distillation with disaggregation (QADD)} trains phase-specific pathways toward a common response objective in one forward-backward pass, supporting shared or separate master weights and prefill-only adaptation to a frozen decoder.
    \item \textbf{Disaggregated quantization} is an umbrella term for three complementary schemes:
    \begin{enumerate}
        \item \textbf{Format disaggregation} combines quantized compute prefill with low-bitwidth weight-only decode. Compared to phase-agnostic computations, this scheme improves accuracy primarily on decode-heavy tasks without increasing weight storage or inference cost.
        \item \textbf{Full disaggregation} trains separate compute-native prefill weights. Compared to weight-only compression, it accelerates prefill while simultaneously boosting accuracy for low-bitwidth decode weights on both prefill-heavy and decode-heavy workloads.
        \item \textbf{Offloaded disaggregated prefill (ODP)} streams prefill weights from SSD through buffers that reuse device memory and overlap loading with compute to avoid additional device memory occupation and, at longer context, hide loading overhead. On Qwen3.8-27B, our \texttt{llama.cpp} implementation delivers a $1.78\times$ time-to-first-token speedup over a weight-only baseline at 8K context.
    \end{enumerate}
    \item \textbf{Prefillers:} training fully-disaggregated NVFP4 prefill checkpoints to augment arbitrary frozen weight-only checkpoints. For 1-bit GGUF compression, we show it more than doubling accuracy over weight-only inference, while simultaneously making prefill faster and memory-efficient via ODP. 
\end{enumerate}


\section{Disaggregated quantization}
\label{sec:dq}

We first introduce QADD (Section~\ref{sec:qad}), then subsequently disaggregate formats, weights and storage (Sections~\ref{sec:prefill_decode}--\ref{sec:offloading}). The measured effect of these schemes on accuracy and inference cost is presented in Section~\ref{sec:exps}.

\subsection{Quantization-aware distillation with disaggregation}
\label{sec:qad}

\emph{Quantization-aware distillation with disaggregation} (QADD) builds on top of quantization-aware distillation (QAD)~\citep{polino2018modelcompressiondistillationquantization,lee2025unifyingblockwiseptqdistillationbased,xin2026quantizationawaredistillationnvfp4inference}. It uses the SFT label mask to propagate the prefill/decode separation from post-training data into the quantized model layers. The mask, normally used for loss masking, now also selects the computational pathway: prompt (user turn) uses prefill, while response (assistant turn) uses decode. Although the distillation loss supervises only response targets, its gradients reach the prefill weights through the prompt keys and values consumed by decode. Both pathways are therefore trained toward the same response objective in one forward-backward pass (Figure~\ref{fig:qadd_training}). Implementation details are provided in Appendix~\ref{app:hyperparams}.

\subsection{Quantization sensitivity depends on the workload}
\label{sec:prefill_decode}

\begin{figure*}[t]
    \centering
    \includegraphics[width=\linewidth]{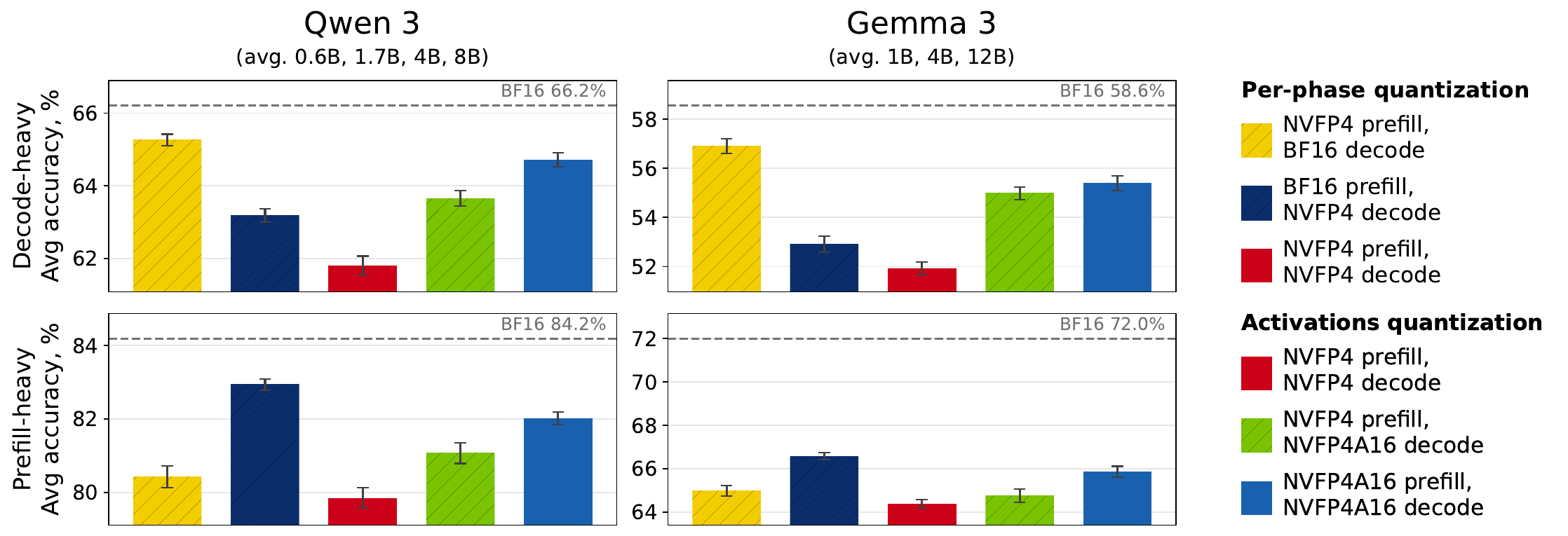}
    \caption{Family-mean accuracy of NVFP4-based formats after quantization-aware distillation on decode-heavy (top) and prefill-heavy (bottom) workloads.  Decode-only NVFP4 quantization is more damaging than prefill-only quantization on decode-heavy tasks; the ordering reverses on prefill-heavy tasks. Format-disaggregated NVFP4 (green) combines NVFP4 prefill with NVFP4A16 decode and improves accuracy over uniform NVFP4 on both workloads without increasing weight storage or prefill cost. On decode-heavy tasks, it approaches the accuracy of NVFP4A16, which uses slower weight-only prefill.}
    \label{fig:bars_phases_both}
\end{figure*}

Before turning to DQ formats, we first establish that different bechmarks interact with quantization of either phase differently, allowing us to monitor phase-specific accuracy effects. We do that by quantizing each phase to NVFP4 in isolation and gauging the effect on two distinct sets of benchmarks: decode-heavy reasoning benchmarks and prefill-heavy benchmarks with long prompts and short answers.

We find that on decode-heavy benchmarks, quantizing decode alone incurs $2$--$4\times$ the accuracy loss of quantizing prefill alone on most models, and up to $7\times$ on Gemma3-1B. On prefill-heavy tasks, prefill-only quantization incurs $1.1$--$4.1\times$ the accuracy loss of decode-only quantization on seven of eight models (Figure~\ref{fig:bars_phases_both}, Figure~\ref{fig:lines_phases_dh}).

With disaggregated quantization, we aim to improve the quality of both stages. Tracking quality on both prefill-heavy and decode-heavy evaluations, verified above, allows us to separate and quantify improvements per stage.

\subsection{Format-disaggregated quantization reduces decode-phase error}
\label{sec:upcast}

NVFP4 quantizes weights and activations alike~\citep{egiazarian2026bridginggappromiseperformance,chen2025intvsfpcomprehensive}, enabling fast prefill; its weight-only variant NVFP4A16 leaves activations unquantized, achieving higher quality but forfeiting the faster prefill computations.

We propose disabling activation quantization only on decode, yielding format-disaggregated NVFP4. It retains original NVFP4's storage and prefill costs while slightly accelerating memory-bound decode by skipping activation quantization (Table~\ref{tab:minimal_comparison}, Appendix~\ref{app:decode_kernels}).

The same approach extends to arbitrary weight encodings and computational formats. Since 2--3-bit quantization has been shown to be Pareto-optimal in size-to-accuracy~\citep{egiazarian2024extremecompressionlargelanguage,liu2025paretoqimprovingscalinglaws,panferov2025queststabletrainingllms}, we also evaluate 2- and 3-bit scalar look-up table (LUT) weight encodings, that we refer to as ``LUT2A16'' and ``LUT3A16'' (Appendix~\ref{app:grids}). To enable native low-precision computations on top of low-bitwidth weights, on-the-fly ``autocast'' re-quantizes them, along with activations, to NVFP4. We refer to these accelerated-compute formats as ``LUT3'' and ``LUT2''. The non-disaggregated scheme applies this autocast in both phases indiscriminately (Figure~\ref{fig:quantized_linear}); format disaggregation restricts it to prefill, combining native NVFP4 prefill computation with compact weight-only LUT decode (Figure~\ref{fig:dq_linear}).

Prefill still uses a re-quantized view of the low-bit decode weights, so its representation remains constrained by their compact encoding, which does not accelerate the NVFP4 computations used by prefill. This motivates giving prefill its own weights.

\subsection{Fully-disaggregated quantization increases prefill-phase capacity}
\label{sec:disag}

\begin{figure}[t]
    \centering
    \includegraphics[width=\linewidth]{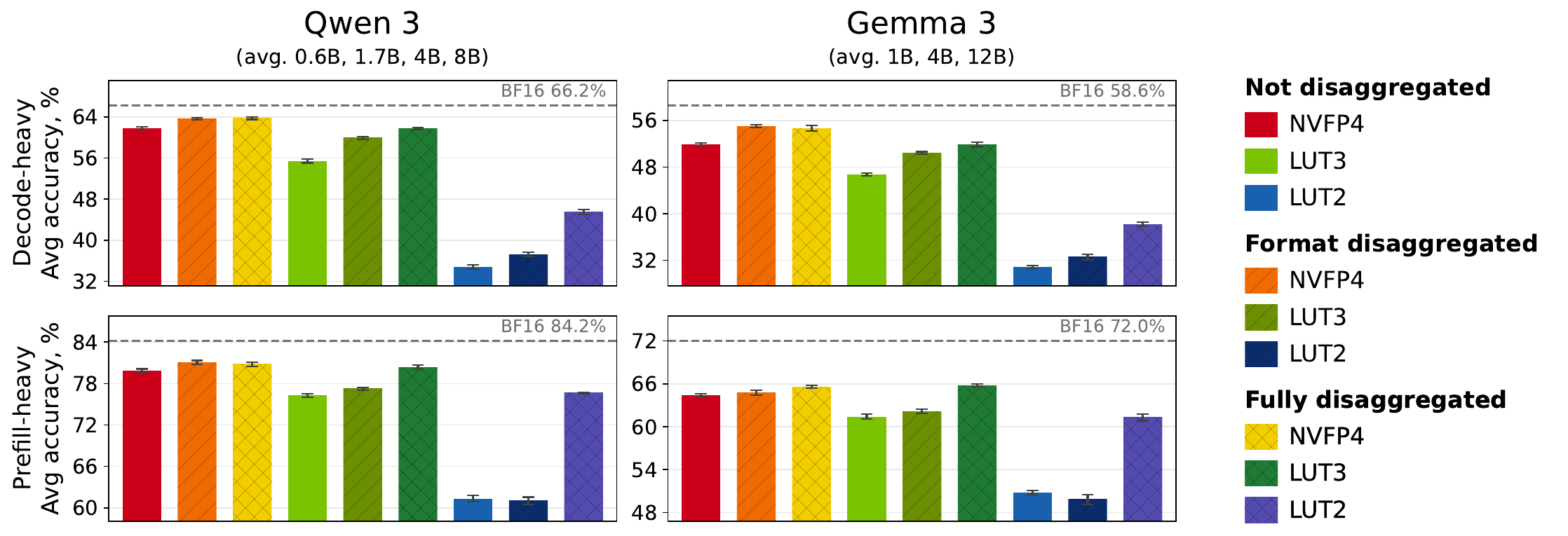}
    \caption{Family-mean accuracy of non-disaggregated, format-disaggregated and fully-disaggregated schemes with $2$--$4$-bit decode weights and NVFP4 prefill compute after quantization-aware distillation. Format disaggregation primarily improves accuracy on decode-heavy tasks (top). Full disaggregation further improves low-bit accuracy on both workload types, with the largest gains over format disaggregation at $2$-bit decode on prefill-heavy tasks.}
    \label{fig:bars_disag_both}
\end{figure}

We propose training separate prefill weights, in a scheme we refer to as ``fully-disaggregated quantization''. Both prefill and decode weights start from the same unquantized model and are optimized together by QADD under their respective formats towards a common response objective, yielding a native NVFP4 prefill checkpoint and a separate weight-only decode checkpoint (Figure~\ref{fig:full_disag_linear}). Each decode format is trained with its own prefill checkpoint, rather than reusing one across bitwidths.

At inference, the prefill checkpoint produces the prompt keys and values at each layer. The decode checkpoint then attends to these representations. The cache retains the original model's layer and head dimensions, so separating the weights does not require any attention modifications. Speed-wise, full disaggregation enjoys the benefits of both NVFP4 prefill computations and weight-only low-bitwidth decode, cleanly combining the best of both worlds.

Full disaggregation mandates storing an additional prefill checkpoint, increasing total storage while preserving the decode weight footprint and speed. Datacenter disaggregation already stores a model instance per phase, but holding both on a single device can be prohibitive. Their residency requirements differ, however. Decode-only weights need high-bandwidth access during generation but are unused during prefill, while prefill weights are needed only during prompt processing and sit idle during decode. We exploit this duality next.

\subsection{Offloaded disaggregated prefill}
\label{sec:offloading}

\begin{figure*}[t]
    \centering
    \vspace{-2mm}
    \begin{subfigure}[b]{0.48\textwidth}
        \centering
        \includegraphics[width=\linewidth]{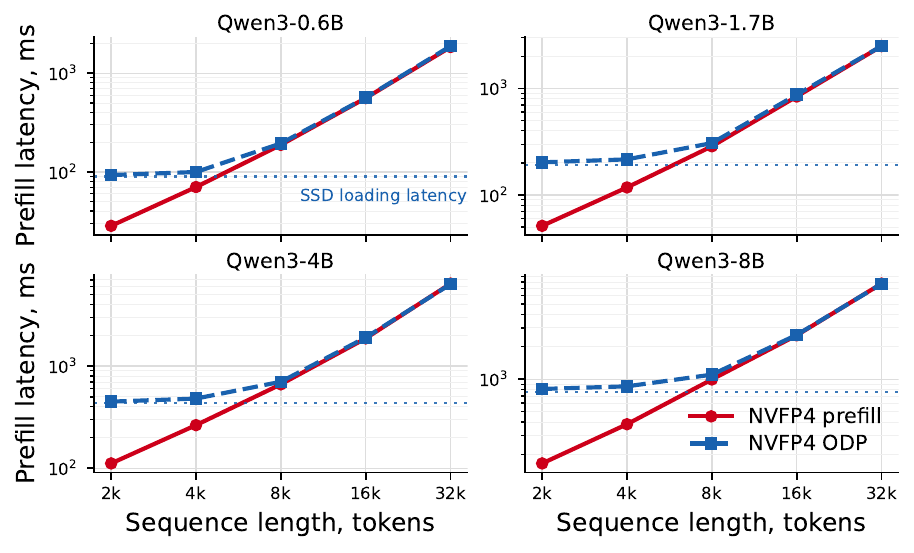}
        \caption{NVFP4 prefill latency with and without ODP. Compute overtakes SSD loading at around 8K context and offloading overhead stays under $5\%$ across Qwen 3 above 16K context.}
        \label{fig:prefill_speed}
    \end{subfigure}
    \hfill
    \begin{subfigure}[b]{0.48\textwidth}
        \centering
        \includegraphics[width=\linewidth]{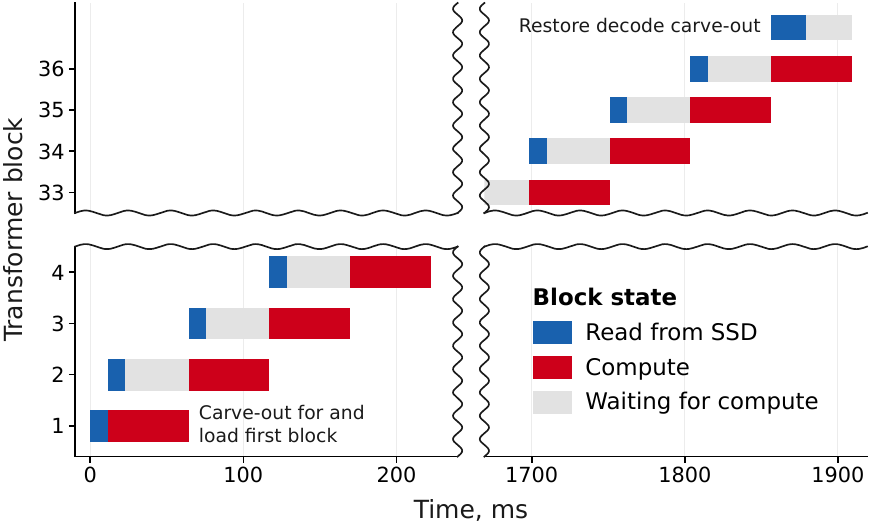}
        \caption{ODP pipeline schematic for Qwen 3 4B at 16K context length in NVFP4, scaled to measured loading and total latency. After a cold start on the first block, subsequent loads overlap with compute.}
        \label{fig:odp_timeline}
    \end{subfigure}
    \vspace{-2mm}
    \caption{Latency effect (a) and pipelining scheme (b) of offloaded disaggregated prefill (ODP).}
    \label{fig:opd}
\end{figure*}

Once a prefill transformer block has produced its outputs, its weights are no longer needed for the rest of the assistant turn, even though its cached keys and values remain in use. \emph{Offloaded disaggregated prefill} (ODP) therefore loads prefill weights from SSD block by block and reuses their device buffers as the context propagates through the network.

Prefill compute grows with context length, while loading the prefill checkpoint from SSD has a fixed cost, so the relative loading overhead decreases as prompts grow. On DGX Spark, compute overtakes loading around 8K context for all Qwen 3 models (Figure~\ref{fig:prefill_speed}). Two device block buffers suffice to overlap compute with loading: while one block processes the prompt, the next is loaded into the other buffer. Longer prompts leave more time for this transfer before the next block is needed, as shown in Figure~\ref{fig:odp_timeline}; short prompts can instead stall on loading. We obtain buffer space by carving-out an equally sized portion of decode weights, unused on prefill, and restoring it before generation. This makes ODP occupy no additional device weight memory at the cost of extra reads.

The same streaming principle, in theory, applies to separate input-processing networks, including encoders in encoder--decoder LLMs and prefill models connected to a decoder through a learned KV-cache adapter~\citep{heo2026crossmodelkvcachetransfer}, replacing full device weight residency with streamed buffers. The same idea, however, does not seamlessly transfer to mixture-of-experts models, as the ratio of compute cost to loading cost grows with the fraction of active parameters, making loading considerably more expensive than compute up to extremely high context lengths. We, therefore, present ODP as mainly a tool for local deployment of dense LLMs.

\subsection{Prefillers for arbitrary weight-only checkpoints}
\label{sec:prequantized}

So far, we have jointly optimized the prefill and decode weights. In practice, however, a suitable quantized decoder may already exist, produced via advanced algorithms over complex encodings~\citep{egiazarian2024extremecompressionlargelanguage,tseng2024quipbetterllmquantization,vanderouderaa2026leechlatticevectorquantization} or released pre-quantized with closed-source or opaque data and algorithms~\citep{gemmateam2026gemma4technicalreport}. As the result, it is possible that existing pre-quantized checkpoints either can't or don't need to be trained during QADD.

We extend full disaggregation to such checkpoints by training only their \emph{prefillers}: prefill models specialized to particular pre-quantized frozen decoders. During training, the decoder uses its dequantized weights without updating them, while gradients propagate through its computations to the prefill pathway. The NVFP4 prefill weights thus adapt to the representations needed by the existing decoder. The decode quantization pipeline can remain a black box: we require its resulting checkpoint, not its training data or optimization algorithm. The resulting model retains the released checkpoint's compressed decode weights and execution pathway while enabling hardware-native NVFP4 prefill. ODP streams the prefiller without increasing device weight residency.

\section{Experimental setup and large-scale validation}
\label{sec:exps}

\begin{table}[t]
\centering
\caption{Cost and accuracy for Qwen 3 and Gemma 3. Speedups and device allocations are for Qwen3-8B and Gemma3-12B. Accuracy (\%) is averaged over 0.6B/1.7B/4B/8B for Qwen 3 and 1B/4B/12B for Gemma 3. DH denotes decode-heavy accuracy over three benchmarks (two reasoning modes for Qwen 3); PH denotes prefill-heavy RULER accuracy over 13 tasks at 4K/8K/16K/32K context. Both use the last five QADD checkpoints (Section~\ref{sec:core_setup}). Prefill is measured at 16K context length on DGX Spark. Decode is measured end-to-end in vLLM as per-token latency at batch one.}
\label{tab:minimal_comparison}
\setlength{\tabcolsep}{2pt}
\small
\input{tables/minimal_comparison.tex}
\end{table}

\subsection{Experimental setup}
\label{sec:core_setup}

\dqpar{Core QADD experiments.} We minimize $\mathrm{KL}(p_\text{teacher}\|p_\text{student})$ against the frozen unquantized teacher on 100M tokens from the T\"{u}lu 3~\citep{lambert2025tulu3pushingfrontiers} SFT corpus. Uniform and disaggregated configurations use the same corpus, token budget and optimization schedule (Appendix~\ref{app:qadd_setup}).

\dqpar{Instruction-tuned models.} DQ requires a logical input/output separation, so we use models that have undergone post-training. Core experiments are performed on Qwen 3~\citep{yang2025qwen3technicalreport} at 0.6B, 1.7B, 4B and 8B parameters, and Gemma 3~\citep{gemmateam2025gemma3technicalreport} at 1B, 4B and 12B parameters. Qwen 3 allows an optional reasoning block during decode, which improves capabilities at the cost of a longer decode phase.

\dqpar{Core QADD benchmarks.} For decode-heavy evaluation, we use the generative versions of GSM8K~\citep{cobbe2021trainingverifierssolvemath}, MATH-500~\citep{hendrycks2021measuringmathematicalproblemsolving} and MMLU-Pro~\citep{wang2024mmluprorobustchallengingmultitask}, with Qwen 3 reasoning both enabled and disabled. For prefill-heavy evaluation, we use RULER~\citep{hsieh2024rulerwhatsrealcontext}, whose long prompts and short answers complement these reasoning workloads. We evaluate its 13 tasks at 4K, 8K, 16K and 32K context lengths, with Qwen 3 reasoning disabled. The reported accuracy summaries use \emph{real} disaggregated serving via vLLM~\citep{kwon2023efficientmemorymanagementlarge} and NIXL. Decode-heavy accuracy is averaged over the three benchmarks (times two modes for Qwen 3). Prefill-heavy scores are averaged over tasks and then context lengths. Per-model accuracy further averages the last five checkpoints of each QADD run, where performance plateaus. A number quoted for a family, such as ``on Qwen 3'', is the unweighted mean over that family's model sizes. Error bars describe two standard deviations over temporal averaging within one training run, not uncertainty across independent runs (Appendix~\ref{app:eval_protocol},\ref{app:breakdown}).

\dqpar{Latency.} We measure end-to-end batch-one per-output-token latency through vLLM and prefill transformer-stack latency with a custom stack built from vLLM kernels. LUT2 and LUT3 decode use custom weight-only kernels. All measurements are performed on DGX Spark (Appendix~\ref{app:speed}).

\subsection{Main QADD-based results}
\label{sec:core_results}

\dqpar{Format-disaggregated quantization.}
Accuracy-wise, on decode-heavy tasks, format disaggregation boosts mean accuracy over the non-disaggregated scheme on all seven models by $1.9$ and $3.1$ for NVFP4, $4.5$ and $3.7$ points for LUT3, $2.5$ and $1.8$ points for LUT2 on Qwen 3 and Gemma 3, respectively (respective improvement for Qwen 3 and Gemma 3 is implied throughout this subsection). On prefill-heavy tasks, however, this decoding-phase optimization has an effect of less than $1.3$ points for all considered model families and formats (Figure~\ref{fig:bars_disag_both}, Table~\ref{tab:minimal_comparison}). 

Speed-wise, on prefill, format disaggregation uses the same accelerated NVFP4 computations as the non-disaggregated scheme, with up to $1.49\times$ and $1.67\times$ speedup over BF16 on Qwen3-8B and Gemma3-12B. On decode, format disaggregation is 2--3\% faster than the non-disaggregated scheme by virtue of skipping activations quantization (Table~\ref{tab:minimal_comparison}, Table~\ref{tab:prefill-per-model}, Figure~\ref{fig:decode_speed}).

That justifies format disaggregation as a plug-in replacement for non-disaggregated inference that boosts accuracy on decode-heavy tasks while retaining or improving all speed and storage costs for single-user serving. It is most useful when device memory is scarce and interactivity of the original model needs to be fully preserved.

\dqpar{Fully-disaggregated quantization and ODP.}
Accuracy-wise, full disaggregation improves both decode-heavy and prefill-heavy performance over non-disaggregated formats. For the former, it yields $6.3$ and $5.2$ points for LUT3, $10.7$ and $7.4$ points for LUT2. For the latter, it gains $4.1$ and $4.3$ points for LUT3, $5.3$ and $10.5$ points for LUT2. For 2--3-bit models, the improvement is noticeable over both non-disaggregated and format-disaggregated serving. For 2-bit models, the gains are so large that fully-disaggregated quantization substantially outperform LUT2A16 weight-only serving, by $4.5$--$12.5$ points, while also delivering faster prefill computations (Figure~\ref{fig:bars_disag_both}, Table~\ref{tab:minimal_comparison}).

Cost-wise, ODP negates the device memory overhead of prefill weights at the cost of constant SSD-bandwidth-bound time-to-first-token and slight prefill latency overhead for longer sequences. At context length above $16$K, offloading increases resident NVFP4 prefill latency by less than $5\%$ on Qwen 3 and $8\%$ on Gemma 3, coming from a cold start on the first block and synchronization logic. At $16$K, the streamed transformer stack remains $1.47\times$ faster than BF16 on Qwen3-8B and $1.58\times$ on Gemma3-12B. In this regime, ODP retains most of the resident NVFP4 prefill speedup without additional device weight memory. Decode speedup remains unchanged relative to format disaggregation (Table~\ref{tab:minimal_comparison}, Table~\ref{tab:prefill-per-model}).

Thus, fully-disaggregated quantization is preferable for both low-concurrency disaggregated serving, when two checkpoints are resident on accelerators anyway and decode is still memory-bound, and for local single-user serving, via ODP. The latter, however, is not as useful for MoE models, and when time-to-first-token (TTFT) on short sequences is critical.

\begin{table}[t]
    \centering
    \caption{Accuracy over text (MMLU-Pro) and image (MMMU-Pro) reasoning benchmarks for SOTA LLMs, as well as gains from 4-bit format disaggregation. Gains highlighted in \textbf{bold} are statistically significant (per-comparison $p<0.05$). $\dagger$: FP8 reference instead of BF16. $\ddagger$: MXFP4 instead of NVFP4, with no BF16 checkpoint available.}
    \label{tab:bigmodels}
    \small
    \setlength{\tabcolsep}{4pt}
    \input{tables/bigmodels.tex}
\end{table}

\dqpar{Prefillers for pre-quantized Qwen3.8-27B.} \label{sec:qwen38}
For the prefillers experiments, we scale our setup to Qwen3.8-27B~\citep{qwen38} --- a dense 27-billion-parameters model released in the summer of 2026. We use eight openly-available pre-quantized GGUF~\citep{llamacpp} checkpoints released by Unsloth~\citep{unsloth}, including vector-quantized formats such as IQ2\_XXS. We evaluate MMLU-Pro~\citep{wang2024mmluprorobustchallengingmultitask} for text reasoning and MMMU-Pro~\citep{yue2025mmmuprorobustmultidisciplinemultimodal}for visual reasoning, with reasoning enabled in both. We report one complete evaluation per format at the end of the $95$M-token QADD training on text-only reasoning traces from the BF16 model on code and math questions (Appendix~\ref{app:qadd27b}). 

Accuracy-wise, training an NVFP4 prefiller improves 1-bit decoder accuracy by $32.5$ points on MMLU-Pro and $35.3$ on MMMU-Pro, more than doubling its weight-only accuracy on both benchmarks. The gains at 2-bit decode are around $7.4$ and $6.3$ points and diminish at higher bitwidths. At 3-bit decode, an NVFP4 prefiller leads to slight performance degradation instead. Figure~\ref{fig:pareto_gsq_rco_both} summarizes the gains across formats and benchmarks.
The QADD corpus contains no multi-modal examples, yet the low-bit accuracy gains transfer strongly to visual reasoning on MMMU-Pro.
For the lowest-bit decoders, learned prefillers also substantially outperform RTN prefill at the same NVFP4 precision, showing that training matters beyond the choice of format (Appendix~\ref{app:interop}).
Appendix~\ref{app:generation_lengths} reports generation-length and truncation measurements.

Speed-wise, adding an NVFP4 prefiller
through our custom \texttt{llama.cpp}~\citep{llamacpp} extension reduces time to first token from $12.27$ to $6.90$ seconds at 8K context, a $1.78\times$ speedup over the weight-only baseline. Across the measured 4K--32K contexts, the speedup ranges from $1.38\times$ to $1.78\times$, while SSD loading makes ODP slower at shorter prompts (Figure~\ref{fig:pareto_gsq_rco_both}).

Building on top of full disaggregation, trained prefillers are most useful in the same serving scenarios as the latter, with the caveat of re-using existing weight-only quantized checkpoints.

\subsection{Large-scale validation with PTQ}
\label{sec:large}

Scaling further up, we apply format disaggregation to eight text and multi-modal models up to $2.8$T parameters. We use NVFP4 PTQ without retraining, testing the most straightforward intervention that is disabling activation quantization only on decode. We evaluate on MMLU-Pro and MMMU-Pro, using paired per-item tests with four evaluation repeats (Appendix~\ref{app:eval_protocol}).
We evaluate Qwen 3.8~\citep{qwen38}, Gemma 4~\citep{gemmateam2026gemma4technicalreport}, Muse Glimmer~\citep{museglimmer}, Nemotron 3~\citep{nvidia2025nvidianemotron3efficient} and Kimi-K3~\citep{kimiteam2026kimik3openfrontier}.

Format disaggregation at 4-bit improves point estimates in 11 of 13 model--benchmark combinations, with 6 significant gains at $\alpha=0.05$ and no significant degradations (Table~\ref{tab:bigmodels}). Significant gains from the most conservative instantiation of disaggregated quantization validate it for SOTA open models with trillions of parameters.

\section{Related work}

Existing work on interaction between inference phases and quantization balances prefill quality against memory-bound decode through phase-specific weight precision~\citep{chen2025progressivemixedprecisiondecodingefficient}, accelerates prefill with low-precision compute while retaining BF16 decode~\citep{lu2026mixquantquantizedprefillingprecise,wei2026hbmneedefficientdisaggregated}, or fine-tunes task-specific prefill modules around a frozen quantized decoder~\citep{woo2026sunsharedusenexttoken}. \citet{forys2026doesdisaggregationpaysimulating} allow prefill and decode to be quantized independently in a disaggregated serving simulator. These works establish the value of phase-specific precision and prefill adaptation. Disaggregated quantization takes this idea further to combine hardware-native prefill with compressed weight-only decode in a unifying approach that can share master weights or maintain separate weights specialized to each phase, supporting both joint optimization and prefill-only adaptation to a frozen decoder.

FlexGen~\citep{sheng2023flexgenhighthroughputgenerativeinference} amortizes weight transfers through large batches. ODP streams the additional prefill checkpoint, amortizing loading over prompt length without increasing device weight residency while decode weights remain resident during generation. This principle may also complement separate prefill networks with learned KV-cache adapters~\citep{heo2026crossmodelkvcachetransfer}, although we do not evaluate that combination.

\section{Conclusion and Limitations}

DQ introduces a phase-aware approach to co-designing quantization formats and model serving. Prefill and decode representations can be chosen for their distinct hardware costs and trained to work together toward a common response objective. This separation also changes local serving: weights needed only during prompt processing can reside on SSD between requests, making room for a specialized prefill model without increasing device weight residency. This phase-aware specialization allows fast prefill, compact decode and accurate responses to coexist for local serving.

Our evaluations cover decode-heavy reasoning and single-turn prefill-heavy tasks, with batch-one decode and prefill-stack timings on DGX Spark and ODP time-to-first-token measurements in \texttt{llama.cpp}. We do not evaluate highly batched performance or multi-turn and agentic behavior. In multi-turn use, cached assistant tokens retain decode-produced KV entries, while rebuilding their cache through prefill can produce different representations for the same token history. Robustness to this cache-policy dependence remains untested.

\paragraph{Released artifacts.}
We release the \href{https://github.com/IST-DASLab/disaggregated-quantization}{codebase} for reproducing our main results, along with a \texttt{llama.cpp} \href{https://github.com/IST-DASLab/disaggregated-llama.cpp}{fork} that adds ODP support, on GitHub. We additionally release the trained Qwen3.8-27B prefillers on the \href{https://huggingface.co/ISTA-DASLab/Qwen3.8-27B-NVFP4-prefiller}{Hugging Face Hub}.

\paragraph{Acknowledgments.}
This research was funded in part by the Austrian Science Fund (FWF) 10.55776/COE12, i.e., the
Bilateral AI Cluster of Excellence, and through generous research support by NVIDIA.
Additionally, we would like
to thank Yoshi Suhara (NVIDIA) for providing and managing the DGX Spark on which the measurements were performed.

\bibliography{iclr2027_conference}
\bibliographystyle{iclr2027_conference}

\appendix

\section{Training and model hyper-parameters}
\label{app:hyperparams}

\subsection{QADD setup}
\label{app:qadd_setup}

Weights and activations are fake-quantized under straight-through estimation~\citep{bengio2013estimatingpropagatinggradientsstochastic}, with FP32 master weights updated by AdamW~\citep{loshchilov2019decoupledweightdecayregularization}.

\dqpar{Hyper-parameters.} Table~\ref{tab:hyper} lists common hyper-parameters for the core Qwen 3 and Gemma 3 QADD runs. The Qwen3.8-27B setup is described in Appendix~\ref{app:qadd27b}. Table~\ref{tab:parallel} lists per-model parallelization hyper-parameters.

\begin{table}[ht]
\centering
\caption{QADD optimization hyperparameters, identical across all formats and models in the core Qwen 3 and Gemma 3 experiments.}
\label{tab:hyper}
\input{tables/hyper.tex}
\end{table}

\begin{table}[ht]
\centering
\caption{Data parallelism (DP), pipeline parallelism (PP) and gradient accumulation setup per model and format class.}
\label{tab:parallel}
\input{tables/parallel.tex}
\end{table}

\dqpar{Phase masking.} The prefill/decode assignment of each input position is read from the SFT labels: positions with an ignored label use prefill, while assistant positions use decode. A per-position mask selects the computational pathway in each quantized linear layer. The loss uses the causal shift, scoring the hidden state at $t$ against the response target at $t+1$, so the final prompt position uses the prefill pathway while predicting the first response token. Prefill weights receive gradients through this boundary prediction and through the prompt keys and values attended to by later response positions. Thus, restricting supervision to response targets still trains both pathways.

\dqpar{Training resources.} Separate-weight configurations maintain optimizer state for both sets of master weights. Table~\ref{tab:parallel} reports hardware allocations. The training corpus, token budget and optimization schedule are not changed between experiments.

\subsection{QADD with a frozen quantized decoder at 27B}
\label{app:qadd27b}

The experiments in Section~\ref{sec:prequantized} use QADD to train an NVFP4 prefiller for each externally quantized decoder. Unlike jointly trained full disaggregation, this setup requires trainable master weights, parameter gradients and optimizer state only for the prefill pathway. The decoder remains in memory during training, using its dequantized BF16 weights, but its parameters are neither updated nor included in the exported prefill checkpoint.

\begin{table}[t]
\centering
\caption{Qwen3.8-27B accuracy (\%) with released weight-only decoders and their NVFP4 prefiller trained by QADD. All eight formats are included, including IQ3\_S and Q3\_K\_XL omitted from Figure~\ref{fig:pareto_gsq_rco_both}. Fully-disaggregated results use step $980$; each score is one complete benchmark evaluation, without checkpoint or repeat averaging. $\Delta$ is the percentage-point change from weight-only, computed before rounding. BF16 is the unquantized reference.}
\label{tab:qadd27b_accuracy}
\small
\setlength{\tabcolsep}{5pt}
\input{tables/qadd27b_accuracy.tex}
\end{table}

\dqpar{Model.} Qwen3.8-27B has $64$ transformer blocks: $48$ use gated linear attention, with full attention in every fourth block. The input embeddings and output projection are untied, and the model includes a vision tower and a multi-modal adapter. Table~\ref{tab:models} lists its dimensions.

\dqpar{Decode checkpoints.} We use eight publicly released Unsloth GGUF checkpoints, spanning nominal $1$--$3$-bit formats: \texttt{IQ1\_S}, \texttt{IQ1\_M}, \texttt{IQ2\_XXS}, \texttt{IQ2\_S}, \texttt{Q2\_K\_XL}, \texttt{IQ3\_XXS}, \texttt{IQ3\_S} and \texttt{Q3\_K\_XL}. Each checkpoint is dequantized to BF16 and frozen, while a separate NVFP4 W4A4 prefill model is trained for each decoder.

\dqpar{Trainable and shared parameters.} We train the prefill copies of the quantized linear weights and text-stack normalization scales, leaving their decode counterparts fixed. Embeddings and the \texttt{lm\_head} are shared between phases and frozen. The narrow linear-attention gate projections \texttt{in\_proj\_a} and \texttt{in\_proj\_b} are also shared and frozen in BF16: they control the recurrent update, and quantizing them destabilized training. The recurrence parameters $A_\text{log}$ and time-step biases likewise remain shared and frozen, preserving the dynamics used by the decode engine. The multi-token-prediction head, conceptually unnecessary for prefill, is not used in this setup and is omitted from export. Exports contain only the prefill checkpoint and the multi-modal adapter.

\dqpar{Corpus.} T\"{u}lu~3 lacks explicit reasoning traces, so the chat template used here inserts an empty thinking block before each answer. For this reasoning-heavy model, we instead distill on \texttt{faunix/Qwen3.8-27B-Distillation-40K}, a public corpus of $40$K traces generated by the same BF16 model as the teacher. Examples requiring tool calls or lacking a nonempty reasoning trace or final answer are discarded ($559$ of $40{,}000$). No image or otherwise multi-modal inputs are present in the mix.

\dqpar{Sequence length and training budget.} The median tokenized sequence and prompt lengths are $3172$ and $111$ tokens, respectively. We therefore increase the sequence-length limit to $8192$ and reduce the global batch size to $32$. Examples are dropped rather than truncated, since truncation can remove important milestones such as closure of the reasoning stage. The filtered corpus contains $32{,}608$ sequences and approximately $95$M tokens, which corresponds to $980$ training steps.

\dqpar{Benchmark overlap.} The training prompts originate from twelve public corpora, so we check their overlap with the evaluation benchmarks. Among the $12{,}032$ MMLU-Pro items, we find one exact prompt match, two matches after normalization, and three with token-shingle Jaccard similarity of at least $0.3$. These matches concern mathematics problems also present in the corpus's math sources. We find no MMMU-Pro prompt matches under these checks. This does not establish that the corpus is generally benchmark-free: its sources include BIG-Bench Hard and OlympiadBench, while its math sources contain MATH items. We therefore do not use those three benchmarks to evaluate this setup.

\subsection{Quantization formats}

Following best practices in grid design and micro-scaling quantization~\citep{blumenberg2025improvingblockwisellmquantization,cook2026adaptiveblockscaleddatatypes,egiazarian2026gridgamespowermultiple}, we tune our scalar LUT quantization scheme to a Gaussian prior and utilize two-level scaling.

\dqpar{NVFP4-esque two-level scaling.} Tensors are split into groups of $16$ elements along the contraction dimension. Each tensor carries one FP32 global scale $s = \max|W| / (6 \cdot 448)$, and each block an FP8-E4M3 scale, so a block's extreme element normalizes to approximately $\pm 6$ and the block scales stay inside E4M3's range. Layers that an inference engine fuses ($q/k/v$ into \texttt{qkv\_proj}, gate/up into \texttt{gate\_up\_proj}) share one global scale, derived  from the group-wise maximum.

\dqpar{NVFP4.} E2M1 elements on the magnitude grid $\{0, 0.5, 1, 1.5, 2, 3, 4, 6\}$ with sign, blocks of 16, FP8-E4M3 block scales and one FP32 per-tensor global scale. W4A4 quantizes activations with a static per-tensor input scale from a running-maximum observer, matching what the serving engine applies at inference.

\dqpar{LUT formats.} \label{app:grids} The weight-only formats use a scalar look-up-table encoding with the same two-level scaling. The grids are asymmetric and contain zero, and the per-block scale now, as opposed to native NVFP4, absorbs the sign of the block's maximum-magnitude element, so after normalization that element is always $+6$ and the value distribution is asymmetric. The grids are

\begin{align*}
\text{LUT3A16 (8 levels):}\quad & \{-4.7038,\; -2.8698,\; -1.3696,\; 0,\; 1.2204,\; 2.5285,\; 4.0473,\; 6\} \\
\text{LUT2A16 (4 levels):}\quad & \{-3.6517,\; 0,\; 2.5227,\; 6\}
\end{align*}

These grids were optimized to minimize the expected quadratic error over $\mathcal{N}(0;1)$ samples scaled with the aforementioned 2-level scaling, constrained to contain $0$ and $+6$.

\dqpar{Storage cost.} A LUT$b$ weight costs $b$ bits per element plus one FP8 scale per 16 elements, i.e. $b + 0.5$ bits per element amortized; NVFP4 costs $4.5$ bits on the same accounting. The fully-disaggregated formats store two checkpoints and therefore pay both. ODP keeps the additional checkpoint on SSD without increasing device weight residency (Section~\ref{sec:offloading}).

The \texttt{lm\_head} is left unquantized in every format.

\subsection{Models}

\begin{table}[ht]
\centering
\caption{Inner shapes of the instruction-tuned models used in this work.}
\label{tab:models}
\input{tables/models.tex}
\end{table}

Per-model hyper-parameters are shown in Table~\ref{tab:models}.

\subsection{Evaluation protocol}
\label{app:eval_protocol}

\dqpar{Decode-heavy QADD results.} We run evaluations with the following parameters: GSM8K (5-shot), MATH-500 (4-shot) and MMLU-Pro (5-shot). For Qwen 3, each is evaluated with reasoning enabled and disabled. Gemma provides no lever to control reasoning and is only evaluated in basic CoT mode. Reported accuracy is the mean over the three benchmarks, and both reasoning modes when applicable, and the last five exported checkpoints of each run, where performance has plateaued; error bars are two standard deviations bootstrapped over that temporal averaging (i.e., steps $1250$, $1500$, $1750$, $2000$, $2250$). These intervals capture late-training checkpoint variation within a single run and do not account for seed or data-order variance. GSM8K is scored with flexible answer extraction, MATH-500 with symbolic verification, and MMLU-Pro with its standard extraction.

\dqpar{Prefill-heavy QADD results.} We evaluate RULER's 13 tasks covering retrieval, multi-hop tracing, aggregation and question answering through the lm-evaluation-harness implementation. Each task uses 500 examples per context length at 4K, 8K, 16K and 32K, with zero-shot prompts, the model's chat template, greedy decoding and task-specific generation budgets of $30$--$128$ tokens (default). Qwen 3 reasoning is disabled to retain the short-answer workload. We evaluate the same QADD checkpoints without benchmark-specific training, using the same disaggregated vLLM serving path as above. For each checkpoint, we average task scores within each context length, then average the four lengths. Reported means use the same five late checkpoints and bootstrap procedure as the decode-heavy results.

\dqpar{Qwen3.8-27B QADD results.} For Section~\ref{sec:prequantized}, we evaluate MMLU-Pro ($12{,}032$ items) with five category-specific CoT examples formatted as alternating user and assistant turns, and MMMU-Pro ($1{,}730$ items) with its zero-shot \texttt{vision} mode. Reasoning is enabled in both benchmarks, with temperature $1.0$, top-$p=0.95$, top-$k=20$, a $32{,}768$-token generation budget and a $65{,}536$-token context limit. MMLU-Pro uses the lm-evaluation-harness answer pattern, counting extraction failures as incorrect; MMMU-Pro uses its official multiple-choice parser, with a fixed seed for its random fallback. Accuracy is computed over all benchmark items, without excluding truncated or unparsed responses.

All arms use vLLM on GB300. The GGUF decode weights are dequantized to BF16 for evaluation; fully-disaggregated arms pair these fixed weights with the trained NVFP4 prefill checkpoint and transfer both attention caches and recurrent state between engines. Figure~\ref{fig:pareto_gsq_rco_both} reports encoded GGUF backbone sizes rather than the allocations of this dequantized evaluation backend. We report one complete evaluation per format and benchmark, without repeat or checkpoint averaging. Fully-disaggregated arms use the final checkpoint at step $980$.

\dqpar{Large and multi-modal PTQ models.} For the models from Section~\ref{sec:large}, we report MMLU-Pro ($12{,}032$ items) and, for the multi-modal models, MMMU-Pro in the \texttt{vision} setting ($1{,}730$ items), scored with each benchmark's standard extraction. All arms are served with vLLM on GB300 nodes; disaggregated arms place the prefill and decode engines on disjoint nodes and transfer the KV cache over NVLink.

Four arms are reported per model. \emph{BF16} is the released checkpoint and \emph{W4A16} is our NVFP4A16 quantization of it. The two \emph{W4A4} arms additionally quantize activations, and differ only in whether that is applied uniformly (\emph{None}) or only to the prefill engine, with decode left weight-only (\emph{Format}). Two models deviate from this scheme: Qwen3.8-2.4T's BF16 is actually FP8 in which it was natively trained and Kimi-K3 is released natively in MXFP4 with no BF16 checkpoint, so it has no BF16 column and its arms are MXFP4 rather than NVFP4. We use the recommended sampling parameters for each model. The model-specific reasoning parser is enabled where one exists. Kimi-K3 is evaluated with thinking enabled at its intermediate effort setting.

\dqpar{Flips analysis for large models.}
We generally follow the per-sample binary answer flips setup of~\citet{kübler2026llmssignificantlyworsestatistical}. We conduct per-item paired tests using four evaluation passes for both MMMU-Pro and MMLU-Pro. With $s^{A}_{i,r}\in\{0,1\}$ the score of method $A$ on item $i$ in pass $r$, we test $\Delta=\frac{1}{n}\sum_i d_i$ where $d_i=\frac{1}{k}\sum_r s^{A}_{i,r}-\frac{1}{k}\sum_r s^{B}_{i,r}$. Under the null that the methods are exchangeable on every item, each $d_i$ is equally likely to flip sign, giving $p=\Pr(|\sum_i\varepsilon_i d_i|\ge|\sum_i d_i|)$ with $\varepsilon_i\sim\mathrm{Unif}\{\pm1\}$. Since $k d_i\in\mathbb{Z}$, the null distribution is a convolution of binomials and is computed exactly rather than sampled. We report per-comparison significance at $\alpha=0.05$, without multiple-testing correction.

\section{Additional ablations}

\subsection{Disaggregation beyond quantized linear layers}
\label{app:full_disag}

\begin{figure}[t]
    \centering
    \includegraphics[width=1.0\linewidth]{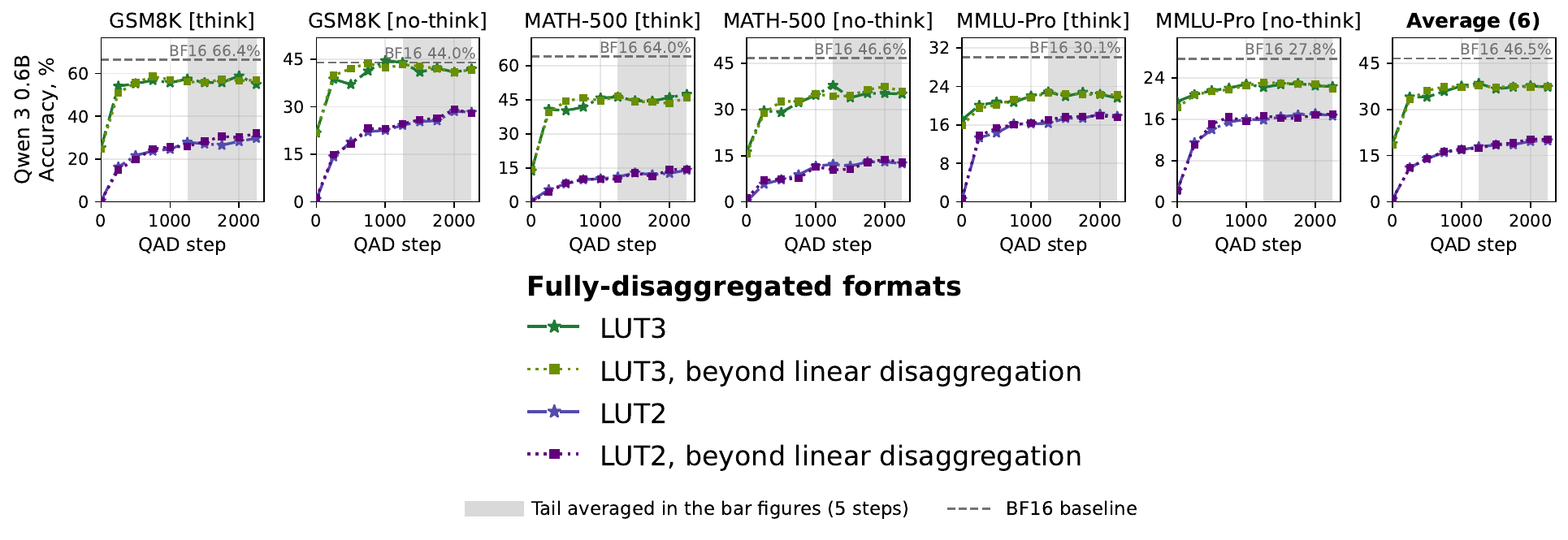}
    \caption{Fully-disaggregated quantization on Qwen3-0.6B with and without separation of the remaining unquantized parameters. This additional separation has no discernible effect on decode-heavy accuracy.}
    \label{fig:lines_non_linear_disag_dh}
\end{figure}

Full disaggregation in Section~\ref{sec:disag} gives the quantized linear layers phase-specific weights while leaving the remaining unquantized parameters shared. We test whether separating these remaining parameters, including embeddings, the output head and layer normalizations, provides further benefit on Qwen3-0.6B, where they account for $26\%$ of all model parameters. This ``beyond linear disaggregation'' has no discernible additional effect on decode-heavy accuracy (Figure~\ref{fig:lines_non_linear_disag_dh}). Both configurations already specialize the quantized linear weights, so this ablation does not distinguish their specialization benefit from that of higher prefill precision.

\subsection{Weight-only quantization results}

In Table~\ref{tab:minimal_comparison}, the reported weight-only quantization schemes are those used in ``LUT3'' and ``LUT2'' formats and described in Appendix~\ref{app:grids}, except naturally without the NVFP4 autocast. Figures~\ref{fig:lines_weight_only_dh} and~\ref{fig:lines_weight_only_ph} break down their decode-heavy and prefill-heavy performance, respectively.

\subsection{Pareto analysis}
\label{app:pareto}

Table~\ref{tab:minimal_comparison} also reports the cost of decode activation quantization. NVFP4 uses the native W4A4 pathway, while format-disaggregated NVFP4 uses NVFP4A16. For LUT2 and LUT3, the non-disaggregated decode timings add an FP4 activation-quantization call before each weight-only linear operation, discarding its output. These timings estimate the overhead of the extra operation, not a complete LUT-to-NVFP4 autocast implementation; they do not measure decode weight re-quantization.

For prefill, the table uses BF16 as the weight-only reference and reuses NVFP4 timings for LUT autocast, excluding weight-conversion overhead.

\begin{figure}[t]
    \centering
    \includegraphics[width=\linewidth]{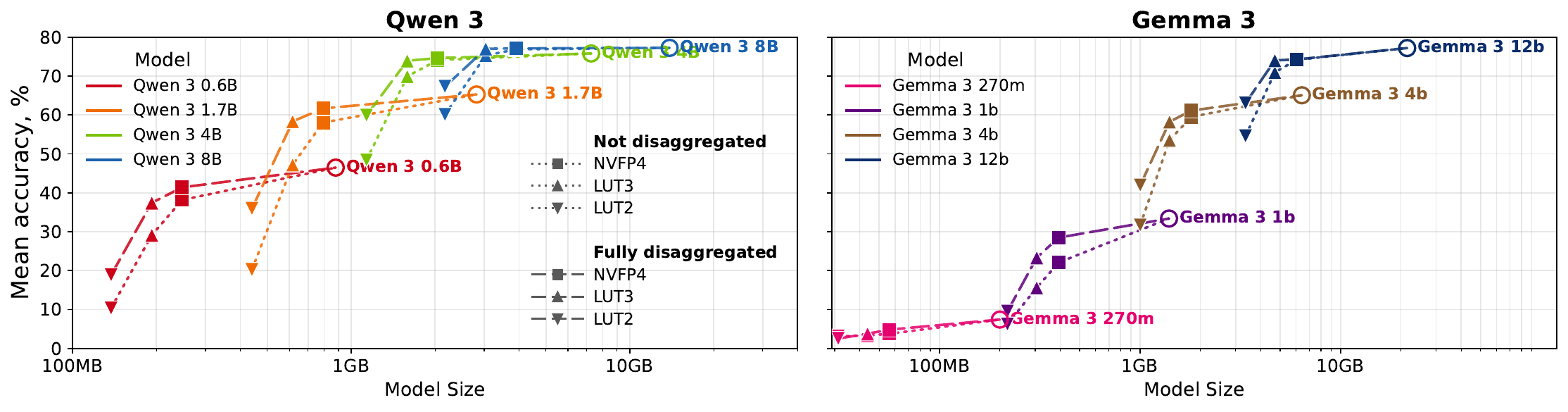}
    \caption{Decode-heavy accuracy of non-disaggregated and fully-disaggregated 2-4-bit quantized models with NVFP4 prefill versus encoded size of the quantized decode weights.}
    \label{fig:pareto_decode_weight_size_dh}
\end{figure}

Figure~\ref{fig:pareto_decode_weight_size_dh} compares decode-heavy accuracy against the encoded size of the quantized decode weights for the Qwen 3 and Gemma 3 families. Full disaggregation improves low-bit accuracy without enlarging these weights, shifting the accuracy--decode-weight-size trade-off. The extra prefill checkpoint and its residency cost are accounted for separately in Table~\ref{tab:minimal_comparison}. Comparisons across model sizes also exclude different amounts of unquantized embedding and head weights.

\subsection{Interoperability of prefillers}
\label{app:interop}

We test how much of the benefit of QADD transfers across decoders and how much depends on the decoder used during training. We compare three frozen Qwen3.8-27B decoders, IQ1\_S, IQ1\_M and IQ2\_XXS, with each of the three step-$980$ prefiller and a common NVFP4 PTQ prefill checkpoint obtained by round-to-nearest (RTN) quantization. All four prefill checkpoints use NVFP4; we identify each QADD-trained prefiller by the decoder used during training. We exchange the prefillers without further training and keep the evaluation protocol of Appendix~\ref{app:eval_protocol} unchanged.

\begin{table}[t]
\centering
\caption{Qwen3.8-27B prefill--decode interoperability, accuracy (\%). Rows fix the decode checkpoint; columns change only the prefill checkpoint. Bold marks the prefiller trained with that decoder, not necessarily the highest score. $\star$ denotes a difference from the bold cell in the same row under a two-sided exact McNemar test ($p<0.05$, without multiple-comparison correction).}
\label{tab:interop}
\small
\setlength{\tabcolsep}{5pt}
\input{tables/interop.tex}
\end{table}

Table~\ref{tab:interop} separates the benefit of training prefill from the benefit of its pairing with a particular decoder. Compared with the common RTN prefill, training-matched prefillers improve MMLU-Pro accuracy by $13.1$, $4.2$ and $1.9$ points for IQ1\_S, IQ1\_M and IQ2\_XXS, respectively. On MMMU-Pro, the corresponding gains are $29.4$ and $9.7$ points for the two lowest-bit formats, while IQ2\_XXS changes little ($-0.3$ points). Thus, at the lowest bitwidths, much of the benefit requires adapting prefill rather than merely replacing its weight-only computations with a separate NVFP4 checkpoint.

The learned checkpoints nevertheless remain partially interoperable. On MMMU-Pro, the IQ1\_S decoder performs best with its own prefiller: accuracy falls from $59.65\%$ to $52.60\%$ with the IQ1\_M prefiller and $47.05\%$ with the IQ2\_XXS prefiller. This ordering is consistent with stronger transfer between closer decode bitwidths in this comparison. It is not universal: the IQ1\_M decoder reaches $62.49\%$ with either its own or the IQ1\_S prefiller, and on MMLU-Pro the IQ1\_S decoder improves from $61.54\%$ to $65.02\%$ when paired with the IQ1\_M prefiller.

\subsection{Generation length with prefillers}
\label{app:generation_lengths}

An unchanged decode checkpoint does not imply unchanged generation cost: the prefiller changes the representations that condition generation and can therefore change response length. Table~\ref{tab:qadd27b_lengths} reports generation lengths and truncation rates for the same Qwen3.8-27B evaluations as Table~\ref{tab:qadd27b_accuracy}, using the final prefill checkpoint at step $980$ and a common $32{,}768$-token generation budget.

\begin{table}[t]
\centering
\caption{Generation length and truncation for Qwen3.8-27B. WO uses the released weight-only checkpoint in both phases; Pref. pairs the same decoder with its trained NVFP4 prefiller at step $980$. Token counts include reasoning and final-answer generation over all benchmark items, including truncated responses. Length statistics are computed from per-item counts and rounded to the nearest token; truncation rates count length-limit terminations. Each arm uses one complete evaluation, with the same quantized checkpoints as Table~\ref{tab:qadd27b_accuracy}.}
\label{tab:qadd27b_lengths}
\small
\setlength{\tabcolsep}{3.5pt}
\input{tables/qadd27b_lengths.tex}
\end{table}

On MMMU-Pro, the largest low-bit accuracy gains coincide with shorter mean responses. IQ1\_S's mean response length falls from $14{,}579$ to $6{,}582$ tokens, a $54.8\%$ reduction, while accuracy rises from $24.4\%$ to $59.7\%$. Its median falls from $10{,}966$ to $2{,}393$ tokens and its truncation rate from $24.57\%$ to $4.16\%$. IQ1\_M likewise uses $14.2\%$ fewer tokens on average while gaining $17.6$ accuracy points. These improvements are therefore not bought with more generated tokens.

The effect depends on the workload and format. On MMLU-Pro, prefillers shorten median generations for seven of eight formats by $3$--$21\%$, but increase mean length by $5$--$63\%$: longer upper tails outweigh the shorter typical responses. On MMMU-Pro, the remaining six formats also increase mean length. Prefill specialization can therefore reduce the total number of decode tokens as well as improve accuracy, but the unchanged decode weights and per-token pathway do not by themselves guarantee lower generation cost.

\section{Speed measurements}
\label{app:speed}

\subsection{Prefill measurements}
\label{app:prefill}

We benchmark on a single NVIDIA GB10 (DGX Spark, \texttt{sm\_121}) with PyTorch~2.13 / CUDA~13.2. These transformer-stack measurements are separate from the \texttt{llama.cpp} TTFT measurements in Figure~\ref{fig:pareto_gsq_rco_both}. For the latter, we integrate the same offloaded NVFP4 prefill pathway into \texttt{llama.cpp}'s prompt processing, without additional kernel optimizations. The baseline uses Unsloth's IQ1\_S checkpoint through native \texttt{llama.cpp} processing. TTFT measurements use three repetitions after one warmup, with prompt caching disabled.

\dqpar{Compute and timing scope.}
BF16 uses cuBLAS through \texttt{torch.nn.Linear}; NVFP4 uses vLLM activation quantization with static global scales and shape-tuned CUTLASS or FlashInfer GEMMs. We fuse compatible projections and activation processing, and use CUDA graphs. Attention backends are selected per layer type (Table~\ref{tab:attn-backend}). The timed region spans the first transformer layer's input through the last layer's output, including attention but excluding embedding lookup, RoPE-table construction, final normalization and the output head.

\begin{table}[t]
\centering
\caption{Attention backend selection for Gemma3-4B: $H=8$, $H_{kv}=4$, $d=256$, window 1024. Latency is in milliseconds per call; bold marks the selected backends.}
\label{tab:attn-backend}
\input{tables/attn-backend.tex}
\end{table}

\dqpar{Offloading protocol.}
For the core Qwen 3 and Gemma 3 models, ODP streams weights from SSD through pinned host buffers into two device slots. Slots borrow decode-weight memory. The benchmark represents the evicted weights by a serialized byte payload equal to the slot allocation and restores it into those buffers. Timing includes the first block's cold load and ends only after restoration completes. Total SSD traffic is the prefill checkpoint plus the carve-out: $3.64+0.20$\,GB for Qwen3-8B and $5.64+0.23$\,GB for Gemma3-12B.

Before each repetition, we request page-cache eviction for both prefill files and the restoration payload. Loading-only references are measured separately for each model and format, including carve-out transfer for ODP.

\begin{table}[t]
\centering
\caption{Prefill transformer-stack speedup over BF16 for the core Qwen 3 and Gemma 3 models on DGX Spark. NVFP4 keeps weights resident; +ODP streams them under the cold-load and carve-out protocol above. BF16 columns give latency in milliseconds. All three arms are measured in the same benchmark session.}
\label{tab:prefill-per-model}
\input{tables/prefill-per-model.tex}
\end{table}

\dqpar{Speedup decomposition.}
Table~\ref{tab:prefill-per-model} reports full-stack results, Table~\ref{tab:nvfp4-breakdown} explains where time is spent within a layer. The $3.0$--$3.4\times$ projection speedups are diluted by attention, normalization, activation processing and quantization. Attention accounts for $37\%$ of NVFP4 device time on Qwen3-8B, which uses global attention throughout, versus $13\%$ on Gemma3-12B, which alternates five sliding-window layers with one global layer. Fused timings do not isolate activation-quantization overhead. Layer wall-clock and profiler timings come from separate runs, so their residual includes measurement variation and is not a direct launch-cost estimate.

\begin{table}[t]
\centering
\caption{Per-component prefill latency at $S{=}16384$. Components sum to profiled device time for one layer; attention is averaged over each model's layer-type mix. Fused operations are grouped: NVFP4 \texttt{norms} includes activation quantization, and \texttt{gate\_up}/\texttt{down} include all chunked GEMM launches. The final row gives full-stack latency divided by the layer count.}
\label{tab:nvfp4-breakdown}
\input{tables/nvfp4-breakdown.tex}
\end{table}

\subsection{Decode kernels}
\label{app:decode_kernels}

\begin{figure}[t]
    \centering
    \begin{subfigure}[b]{0.48\textwidth}
        \centering
        \includegraphics[width=\linewidth]{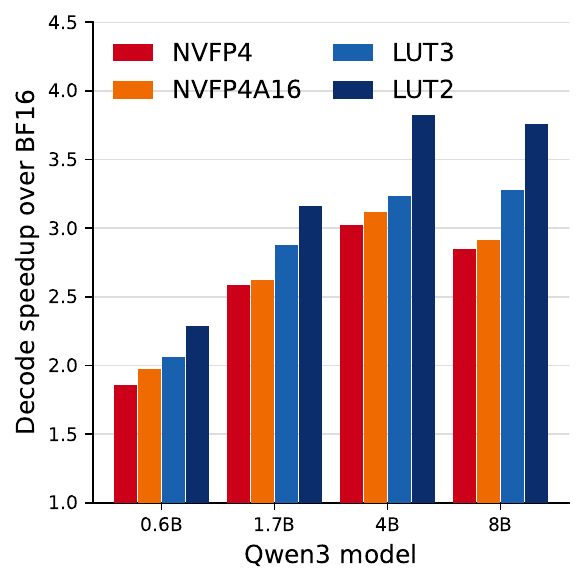}
        \caption{Qwen 3}
    \end{subfigure}
    \hfill
    \begin{subfigure}[b]{0.48\textwidth}
        \centering
        \includegraphics[width=\linewidth]{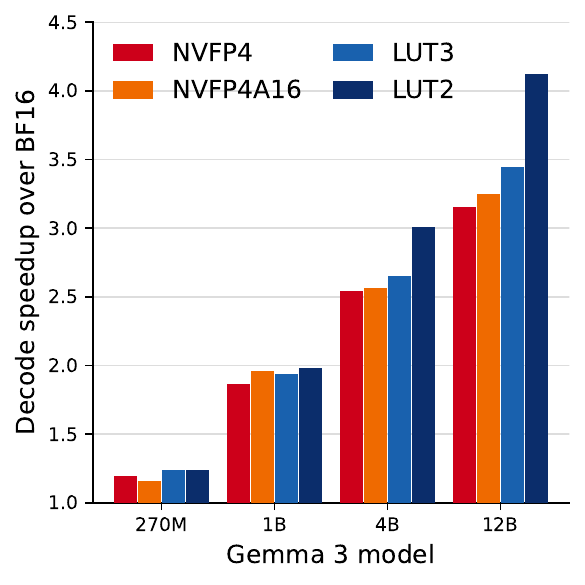}
        \caption{Gemma 3}
    \end{subfigure}
    \caption{End-to-end single-user decode speedup over dense BF16, measured in vLLM on DGX Spark. The speedup is whole-model per-output-token latency, so it includes attention, normalisation and the BF16 language-modelling head, none of which any of these formats accelerates.}
    \label{fig:decode_speed}
\end{figure}

We back up our claims of decode speed depending on the degree of compression by benchmarking memory-bound GEMV kernels when used for LLM decoding in vLLM. Figure~\ref{fig:decode_speed} shows real speedups measured on DGX Spark for NVFP4 and NVFP4A16, natively shipped with vLLM, as well as LUT3 and LUT2 kernels that we implemented and integrated.

\dqpar{Protocol.} All numbers are end-to-end vLLM generations at batch one. The same 168-token prompt is generated to $n_1{=}8$ and $n_2{=}128$ output tokens, and the per-output-token latency is estimated as $(t_2-t_1)/(n_2-n_1)$, which removes the common prefill contribution without relying on engine-internal metrics. Each point is the median of three repetitions after a warm-up generation at that shape. Prefix caching is disabled. CUDA graphs are enabled.

\dqpar{Kernel specification.} Decode at batch one is bandwidth-bound: the arithmetic is a matrix-vector product, so time is set almost entirely by the bytes of weight pulled per token. The four formats differ mainly in that quantity --- BF16 at $16$, NVFP4 at $4.5$, LUT3 at $3.5$ and LUT2 at $2.5$ bits per weight. NVFP4 is vLLM's native W4A4 CUTLASS path, which quantizes the activation before every projection and issues an FP4$\times$FP4 GEMM. NVFP4A16 keeps activations in BF16 and uses a Marlin-like~\citep{frantar2024marlinmixedprecisionautoregressiveparallel} mixed-precision kernel. Our LUT3 and LUT2 kernels store weights bit-planed as \texttt{int32} over $(N, b, K/32)$ with signed \texttt{e4m3} block scales every 16 elements and one FP32 global scale per tensor, and reconstruct values by indexing a $2^b$-entry lookup table in shared memory. This lookup is implemented in software on DGX Spark and does not require native LUT arithmetic support.

Notably, format-disaggregated NVFP4 (identical to NVFP4A16 on decode) outperforms NVFP4 by skipping activation quantization and using a kernel tailored to the weight-only matrix-vector product. The speedup over BF16 increases from $2.86\times$ to $2.93\times$ on Qwen3-8B and from $3.18\times$ to $3.27\times$ on Gemma3-12B.

\dqpar{Speedup dilution.} LUT2 moves $6.4\times$ fewer weight bits than BF16 yet delivers $3.82\times$ and $4.15\times$ speedups, respectively. Two effects account for the gap. Firstly, the output head stays in BF16 in every arm, and on the smaller models its weights account for a substantial fraction of the bytes moved per token. Secondly, embedding lookup, attention, the normalizations and the residual adds are identical work in every arm, so they dilute the speedup. The same principle applies to unaccelerated operations in prefill (Table~\ref{tab:nvfp4-breakdown}). On its own, the LUT2 kernel reaches $96\%$ of the bandwidth its weight traffic permits.

\section{Full evaluation results}
\label{app:breakdown}

\begin{figure}[t]
    \centering
    \includegraphics[width=1.0\linewidth]{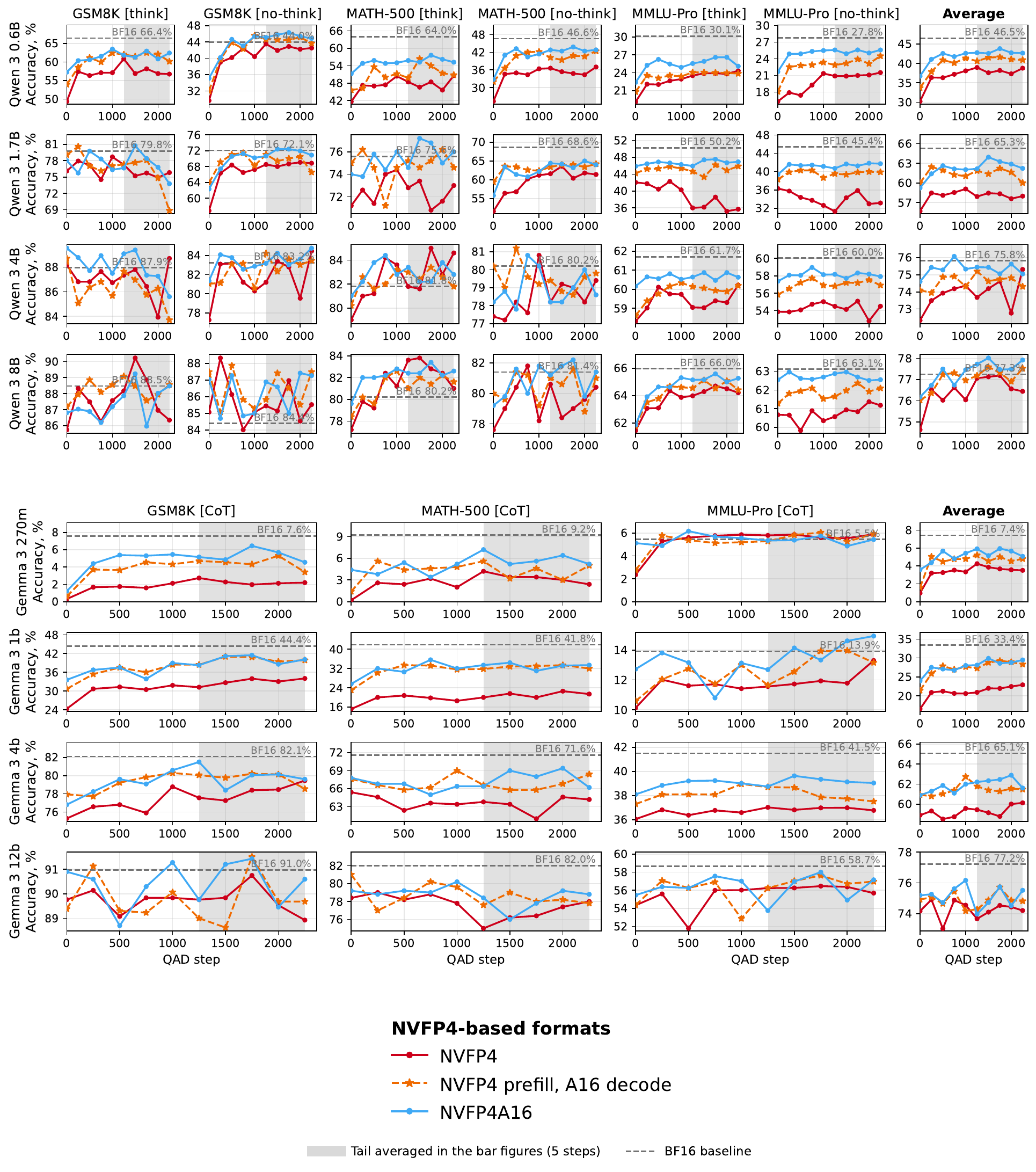}
    \caption{Breakdown of Figure~\ref{fig:bars_phases_both} by benchmark and QADD step.}
    \label{fig:lines_activation_quantization_dh}
\end{figure}

\begin{figure}[t]
    \centering
    \includegraphics[width=1.0\linewidth]{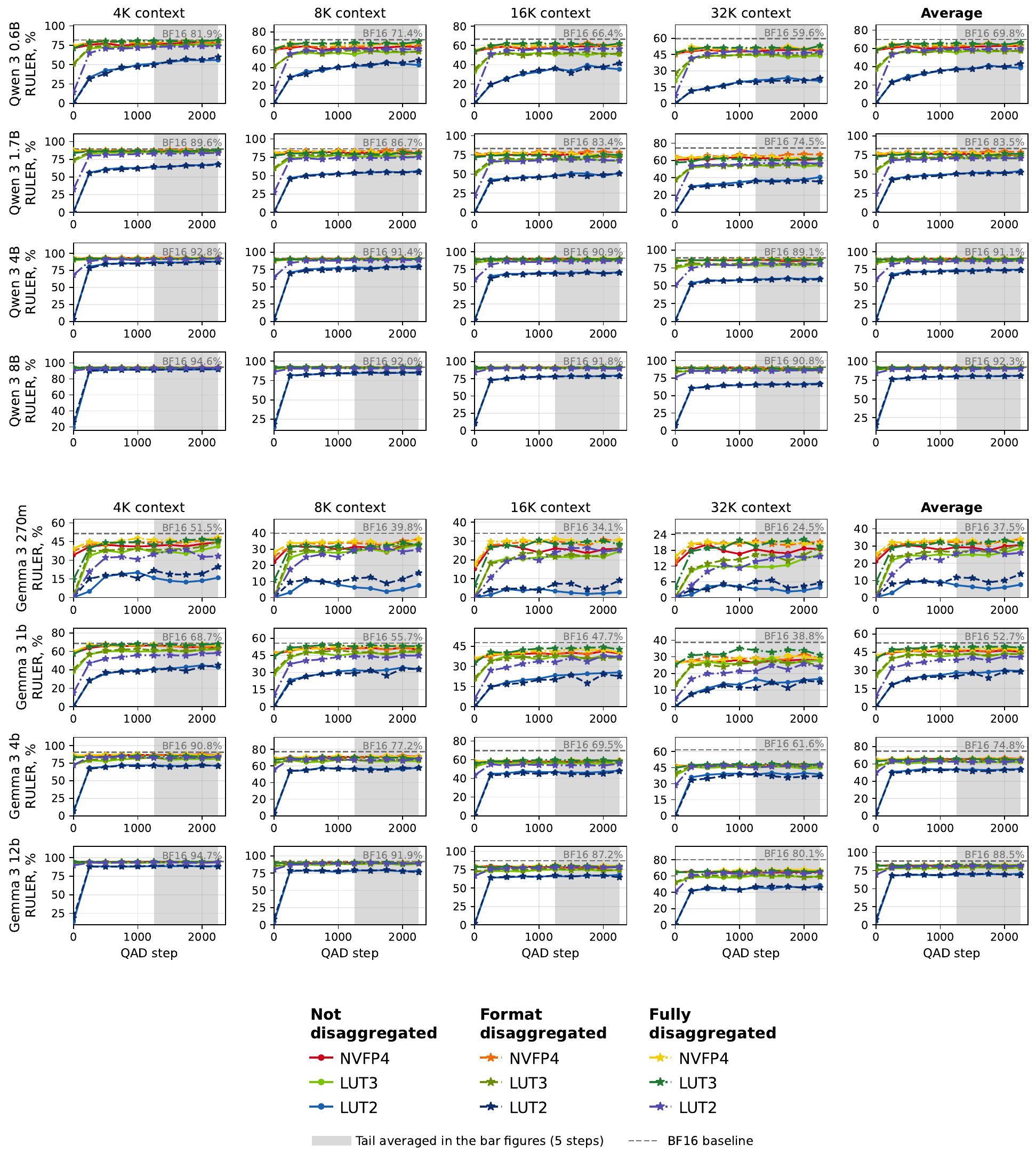}
    \caption{Breakdown of Figure~\ref{fig:bars_disag_both} by context length and QADD step for prefill-heavy tasks.}
    \label{fig:lines_disag_ph}
\end{figure}

\begin{figure}[t]
    \centering
    \includegraphics[width=1.0\linewidth]{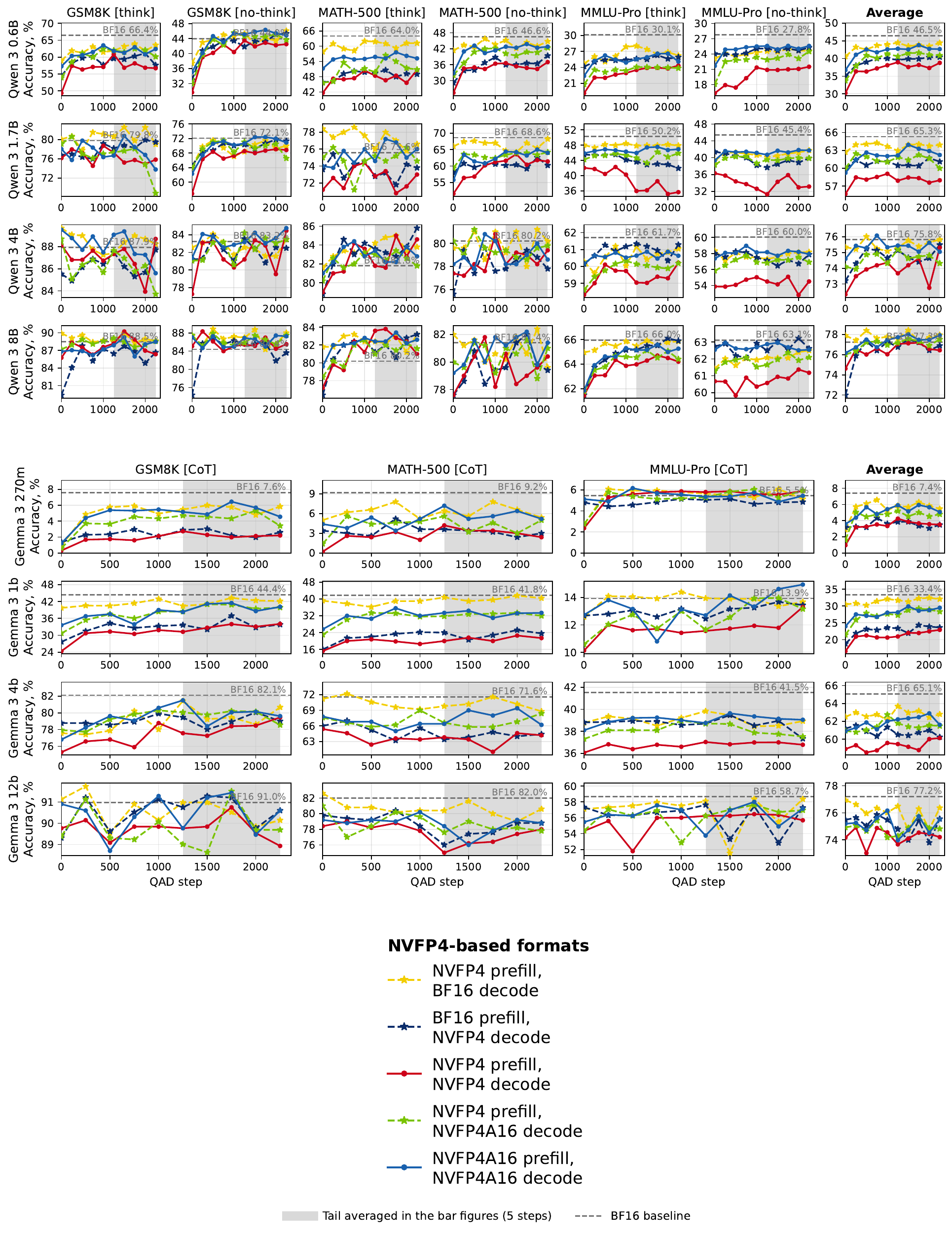}
    \caption{Breakdown of Figure~\ref{fig:bars_phases_both} by benchmark and QADD step for decode-heavy tasks.}
    \label{fig:lines_phases_dh}
\end{figure}

\begin{figure}[t]
    \centering
    \includegraphics[width=1.0\linewidth]{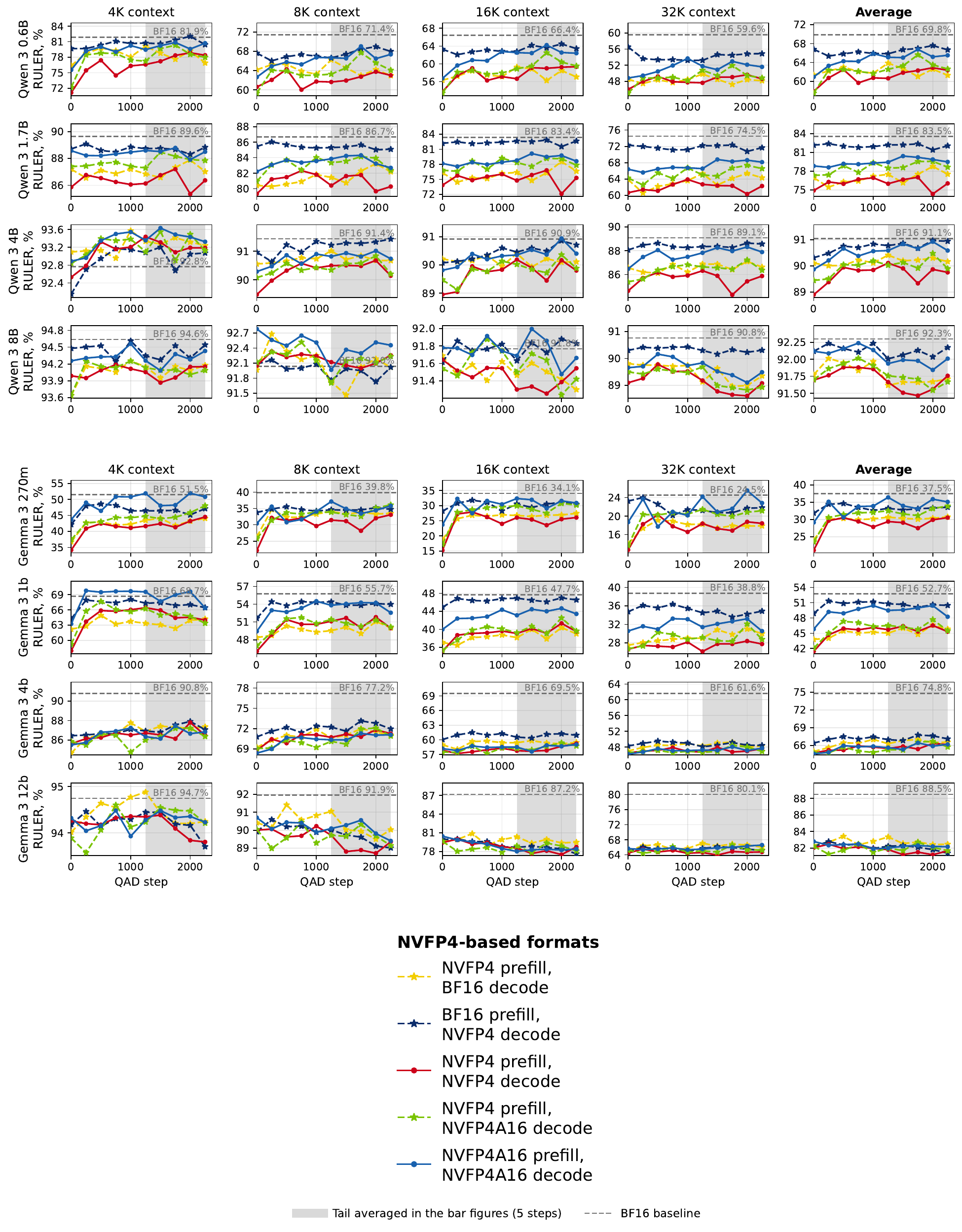}
    \caption{Breakdown of Figure~\ref{fig:bars_phases_both} by context length and QADD step for prefill-heavy tasks.}
    \label{fig:lines_phases_ph}
\end{figure}

\begin{figure}[t]
    \centering
    \includegraphics[width=1.0\linewidth]{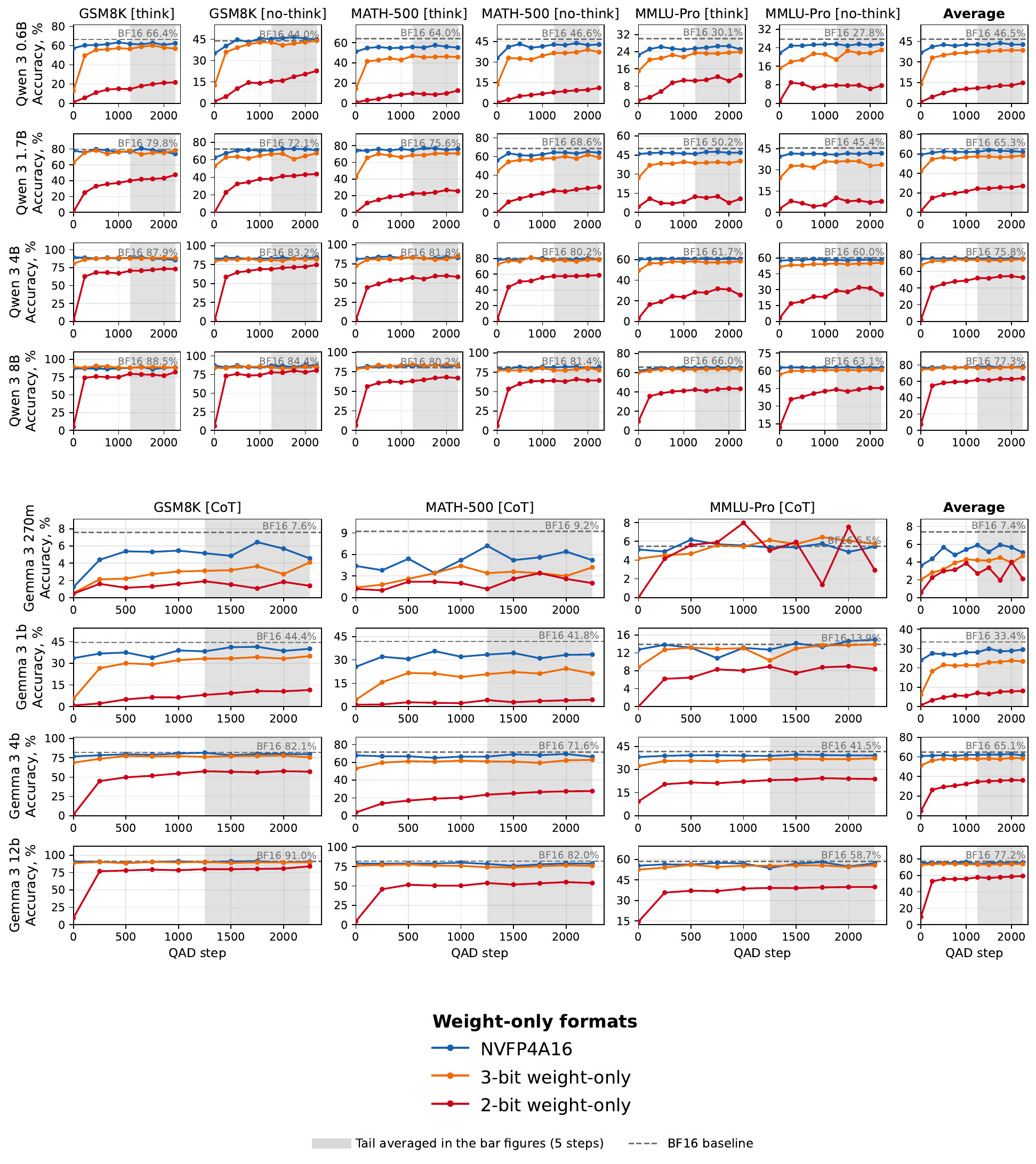}
    \caption{Performance breakdown for weight-only quantization formats by benchmark and QADD step.}
    \label{fig:lines_weight_only_dh}
\end{figure}

\begin{figure}[t]
    \centering
    \includegraphics[width=1.0\linewidth]{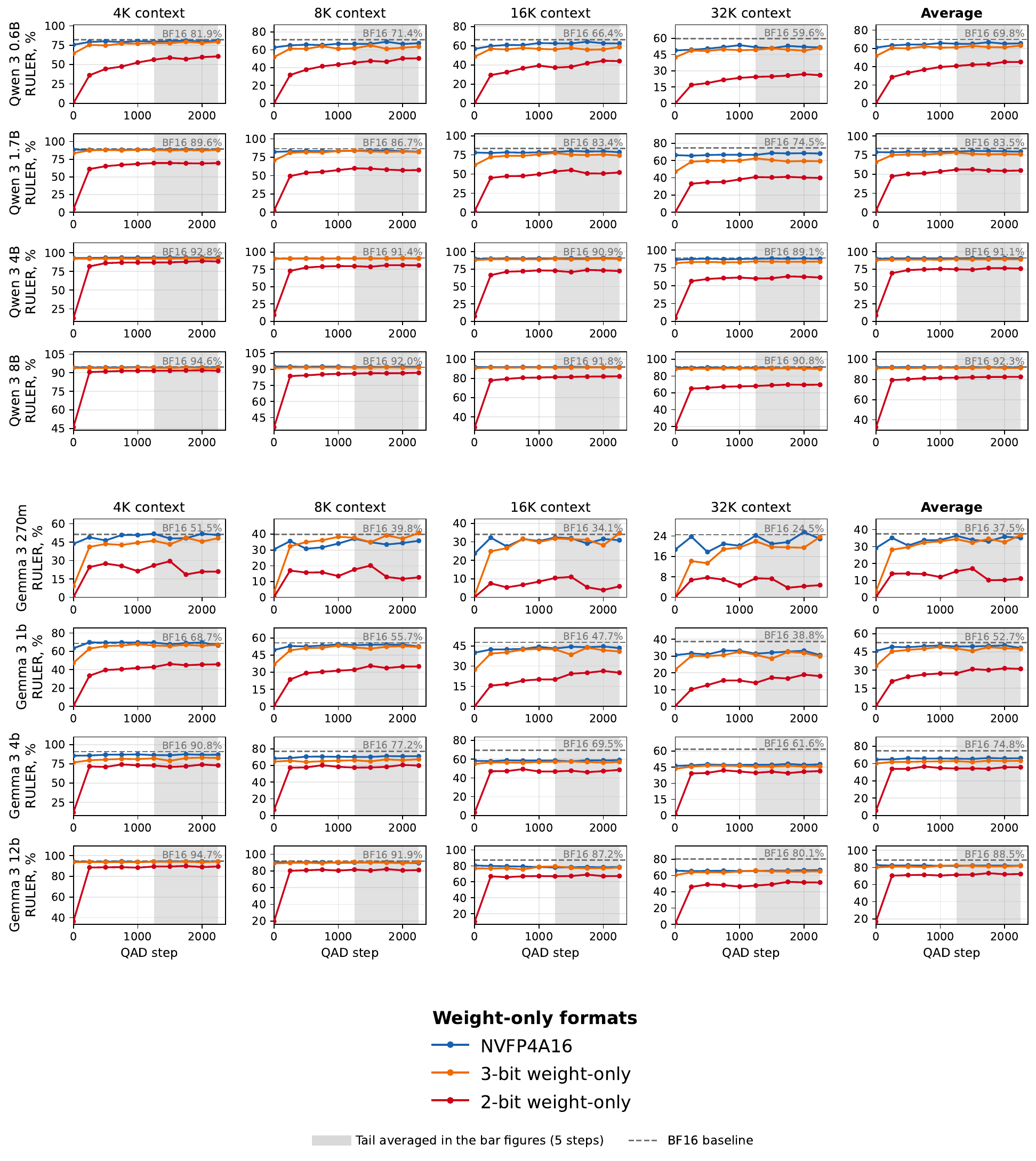}
    \caption{Performance breakdown for weight-only quantization formats on prefill-heavy tasks by context length and QADD step. The average excludes 64K evaluations.}
    \label{fig:lines_weight_only_ph}
\end{figure}

Figures~\ref{fig:lines_activation_quantization_dh},~\ref{fig:lines_phases_dh} and~\ref{fig:lines_weight_only_dh} break down decode-heavy QADD accuracy recovery by benchmark and training step. Figures~\ref{fig:lines_phases_ph},~\ref{fig:lines_disag_ph} and~\ref{fig:lines_weight_only_ph} provide the corresponding prefill-heavy breakdowns by context length and training step.

\end{document}

%% file: math_commands.tex
\usepackage{amsmath,amsfonts,bm}

\def\eqref#1{equation~\ref{#1}}

\def\1{\bm{1}}

\DeclareMathAlphabet{\mathsfit}{\encodingdefault}{\sfdefault}{m}{sl}
\SetMathAlphabet{\mathsfit}{bold}{\encodingdefault}{\sfdefault}{bx}{n}



%% file: tables/minimal_comparison.tex
\begin{tabular}{l|ccccc|ccccc}
\toprule
\multirow{2}{*}{Format} & \multicolumn{5}{c|}{Qwen 3} & \multicolumn{5}{c}{Gemma 3} \\
  & \makecell{Prefill\\speedup} & \makecell{Decode\\speedup} & \makecell{Device\\GB} & \makecell{Acc.\\DH} & \makecell{Acc.\\PH} & \makecell{Prefill\\speedup} & \makecell{Decode\\speedup} & \makecell{Device\\GB} & \makecell{Acc.\\DH} & \makecell{Acc.\\PH} \\ \midrule
BF16 & 1.00x & 1.00x & 16.38 & 66.2 & 84.2 & 1.00x & 1.00x & 23.53 & 58.6 & 72.0 \\\midrule
NVFP4A16 & 1.00x & 2.93x & 6.40 & 64.7 & 82.0 & 1.00x & 3.27x & 8.07 & 55.4 & 65.9 \\
NVFP4 & 1.49x & 2.86x & 6.40 & 61.8 & 79.8 & 1.67x & 3.18x & 8.07 & 51.9 & 64.4 \\
+Format disagg. & 1.49x & 2.93x & 6.40 & 63.7 & 81.1 & 1.67x & 3.27x & 8.07 & 55.0 & 64.7 \\\midrule
3-bit weight-only & 1.00x & 3.33x & 5.53 & 61.5 & 79.8 & 1.00x & 3.44x & 6.72 & 51.6 & 64.1 \\
LUT3 & 1.49x & 3.23x & 5.53 & 55.4 & 76.3 & 1.67x & 3.37x & 6.72 & 46.7 & 61.4 \\
+Format disagg. & 1.49x & 3.33x & 5.53 & 59.9 & 77.2 & 1.67x & 3.44x & 6.72 & 50.4 & 62.1 \\
+Full disagg. & 1.49x & 3.33x & 9.44 & 61.7 & 80.4 & 1.67x & 3.44x & 12.77 & 51.9 & 65.7 \\
+ODP & 1.47x & 3.33x & 5.53 & 61.7 & 80.4 & 1.58x & 3.44x & 6.72 & 51.9 & 65.7 \\\midrule
2-bit weight-only & 1.00x & 3.82x & 4.66 & 38.4 & 64.1 & 1.00x & 4.15x & 5.38 & 33.7 & 52.4 \\
LUT2 & 1.49x & 3.73x & 4.66 & 34.8 & 61.3 & 1.67x & 4.03x & 5.38 & 30.8 & 50.8 \\
+Format disagg. & 1.49x & 3.82x & 4.66 & 37.2 & 61.0 & 1.67x & 4.15x & 5.38 & 32.6 & 49.8 \\
+Full disagg. & 1.49x & 3.82x & 8.57 & 45.5 & 76.6 & 1.67x & 4.15x & 11.43 & 38.2 & 61.3 \\
+ODP & 1.47x & 3.82x & 4.66 & 45.5 & 76.6 & 1.58x & 4.15x & 5.38 & 38.2 & 61.3 \\
\bottomrule
\end{tabular}

%% file: tables/bigmodels.tex
\begin{tabular}{l|cc | ccc|cc | ccc}
    \toprule
    \multirow{3}{*}{Model} & \multicolumn{5}{c|}{MMLU-Pro} & \multicolumn{5}{c}{MMMU-Pro} \\
    \cmidrule(lr){2-6}\cmidrule(lr){7-11}
     & \multirow{2}{*}{BF16} & \multirow{2}{*}{W4A16} & \multicolumn{3}{c|}{W4A4 disaggregation} & \multirow{2}{*}{BF16} & \multirow{2}{*}{W4A16} & \multicolumn{3}{c}{W4A4 disaggregation} \\
     &  &  & None & Format & $\Delta$ &  &  & None & Format & $\Delta$ \\
    \midrule
    Qwen3.8-27B        & 84.58  & 82.33  & 81.47  & 81.83  & \textbf{+0.35} & 74.78  & 71.16  & 69.84  & 70.46  & +0.62  \\
    Gemma-4-31B        & 84.92  & 84.51  & 84.07  & 84.17  & +0.10  & 66.84  & 65.36  & 64.81  & 65.94  & \textbf{+1.13} \\
    Muse-Glimmer-30B   & 75.43  & 76.80  & 75.12  & 76.04  & \textbf{+0.92} & 72.92  & 71.81  & 70.38  & 71.34  & \textbf{+0.97} \\\midrule
    Gemma-4-26B        & 82.40  & 81.01  & 79.74  & 80.52  & \textbf{+0.78} & 63.66  & 60.95  & 59.05  & 59.68  & +0.64  \\
    Nemotron-3-120B    & 83.01  & 82.78  & 82.66  & 82.68  & +0.02  & -      & -      & -      & -      & -      \\
    Nemotron-3-550B    & 86.71  & 86.47  & 86.51  & 86.37  & -0.14  & -      & -      & -      & -      & -      \\
    Qwen3.8-2.4T$^{\dagger}$ & 88.99  & 84.73  & 84.28  & 84.68  & \textbf{+0.40} & -      & -      & -      & -      & -      \\
    Kimi-K3-2.8T$^{\ddagger}$ & -      & 85.94  & 85.78  & 85.57  & -0.21  & -      & 79.93  & 79.18  & 79.71  & +0.53  \\
    \bottomrule
    \end{tabular}

%% file: tables/hyper.tex
\begin{tabular}{ll}
\toprule
Objective & $\mathrm{KL}(p_\text{teacher}\|p_\text{student})$, teacher frozen in BF16 \\
Corpus & T\"{u}lu 3 SFT, 100M non-padding tokens, mean sequence length 638 \\
Sequence length & 2048 \\
Global batch size & 64 sequences \\
Steps & $\approx$2450 \\
Optimizer & AdamW, $\beta=(0.9,0.95)$, $\epsilon=10^{-8}$ \\
Learning rate & $3\times10^{-6}$, constant \\
Warmup & 100 steps \\
Weight decay & 0.1 \\
Gradient clipping & 1.0, per-DDP-rank \\
Master weights & FP32; straight-through estimator to the quantized weight \\
Compute precision & BF16 autocast; FP32 residual stream \\
Parallelism & DDP, ZeRO-2, pipeline parallelism \\
\bottomrule
\end{tabular}

%% file: tables/parallel.tex
\begin{tabular}{lccccc}
\toprule
Model / format & B300 GPUs & PP & DP & Micro-batch & Accum. \\
\midrule
Qwen 3 0.6B / 1.7B / 4B, all formats & 8 & 1 & 8 & 8 & 1 \\
Qwen3-8B, single-master formats & 8 & 1 & 8 & 8 & 1 \\
Qwen3-8B, fully-disaggregated formats & 16 & 2 & 8 & 8 & 1 \\
Gemma 3 270m / 1b / 4b, all formats & 8 & 1 & 8 & 8 & 1 \\
Gemma-3-12b, single-master formats & 16 & 1 & 16 & 4 & 1 \\
Gemma-3-12b, fully-disaggregated formats & 32 & 2 & 16 & 4 & 1 \\
Qwen3.8-27B, frozen-decode & 64 & 2 & 32 & 1 & 1 \\
\bottomrule
\end{tabular}

%% file: tables/qadd27b_accuracy.tex
\begin{tabular}{l|rrr|rrr}
\toprule
\multirow{2}{*}{Decode format} & \multicolumn{3}{c|}{MMLU-Pro} & \multicolumn{3}{c}{MMMU-Pro} \\
 & Weight-only & Full disagg. & $\Delta$ & Weight-only & Full disagg. & $\Delta$ \\
\midrule
BF16 & 84.62 & -- & -- & 75.26 & -- & -- \\
\midrule
IQ1\_S & 29.04 & 61.54 & +32.50 & 24.39 & 59.65 & +35.26 \\
IQ1\_M & 52.88 & 72.59 & +19.71 & 44.86 & 62.49 & +17.63 \\
IQ2\_XXS & 70.50 & 77.93 & +7.42 & 59.60 & 65.90 & +6.30 \\
IQ2\_S & 77.96 & 80.76 & +2.80 & 68.50 & 69.36 & +0.87 \\
Q2\_K\_XL & 82.16 & 82.35 & +0.18 & 72.31 & 71.68 & -0.64 \\
IQ3\_XXS & 82.11 & 82.84 & +0.72 & 73.41 & 71.45 & -1.97 \\
IQ3\_S & 83.65 & 83.02 & -0.63 & 74.74 & 71.97 & -2.77 \\
Q3\_K\_XL & 84.25 & 83.64 & -0.62 & 74.28 & 72.31 & -1.97 \\
\bottomrule
\end{tabular}

%% file: tables/models.tex
\begin{tabular}{lccccc}
\toprule
Model & Layers & $d_\text{model}$ & $d_\text{ffn}$ & Vocabulary & Tied embeddings \\
\midrule
Qwen3-0.6B & 28 & 1024 & 3072 & 151936 & yes \\
Qwen3-1.7B & 28 & 2048 & 6144 & 151936 & yes \\
Qwen3-4B & 36 & 2560 & 9728 & 151936 & yes \\
Qwen3-8B & 36 & 4096 & 12288 & 151936 & no \\
\midrule
Gemma-3-270m & 18 & 640 & 2048 & 262208 & yes \\
Gemma-3-1b & 26 & 1152 & 6912 & 262208 & yes \\
Gemma-3-4b & 34 & 2560 & 10240 & 262208 & yes \\
Gemma-3-12b & 48 & 3840 & 15360 & 262208 & yes \\
\midrule
Qwen3.8-27B & 64 & 5120 & 17408 & 248320 & no \\
\bottomrule
\end{tabular}

%% file: tables/interop.tex
\begin{tabular}{l|rrrr}
\toprule
\multicolumn{5}{c}{\textbf{MMLU-Pro} ($n=12032$)} \\
\multirow{2}{*}{Frozen decode} & \multicolumn{4}{c}{Prefill checkpoint (all NVFP4)} \\
 & PTQ (RTN) & QADD IQ1\_S & QADD IQ1\_M & QADD IQ2\_XXS \\
\midrule
IQ1\_S & 48.39$^{\star}$ & \textbf{61.54} & 65.02$^{\star}$ & 62.59$^{\star}$ \\
IQ1\_M & 68.37$^{\star}$ & 69.62$^{\star}$ & \textbf{72.59} & 72.85 \\
IQ2\_XXS & 76.06$^{\star}$ & 74.27$^{\star}$ & 77.44 & \textbf{77.93} \\
\midrule
\multicolumn{5}{c}{\textbf{MMMU-Pro} ($n=1730$)} \\
\multirow{2}{*}{Frozen decode} & \multicolumn{4}{c}{Prefill checkpoint (all NVFP4)} \\
 & PTQ (RTN) & QADD IQ1\_S & QADD IQ1\_M & QADD IQ2\_XXS \\
\midrule
IQ1\_S & 30.29$^{\star}$ & \textbf{59.65} & 52.60$^{\star}$ & 47.05$^{\star}$ \\
IQ1\_M & 52.83$^{\star}$ & 62.49 & \textbf{62.49} & 61.50 \\
IQ2\_XXS & 66.18 & 67.11 & 66.30 & \textbf{65.90} \\
\bottomrule
\end{tabular}

%% file: tables/qadd27b_lengths.tex
\begin{tabular}{l|rr|rr|rr|rr}
\toprule
\multicolumn{9}{c}{\textbf{MMLU-Pro}} \\
\multirow{2}{*}{Decode format} & \multicolumn{2}{c|}{Mean tokens} & \multicolumn{2}{c|}{Median tokens} & \multicolumn{2}{c|}{95th percentile} & \multicolumn{2}{c}{Truncated (\%)} \\
 & WO & Pref. & WO & Pref. & WO & Pref. & WO & Pref. \\
\midrule
IQ1\_S & 3964 & 4260 & 928 & 778 & 18952 & 26677 & 2.46 & 3.34 \\
IQ1\_M & 3312 & 3612 & 883 & 699 & 16910 & 23550 & 2.42 & 3.39 \\
IQ2\_XXS & 2394 & 3897 & 746 & 782 & 11191 & 23999 & 1.25 & 3.39 \\
IQ2\_S & 1706 & 2766 & 774 & 671 & 6352 & 14429 & 0.27 & 1.53 \\
Q2\_K\_XL & 2647 & 2787 & 799 & 671 & 13092 & 14721 & 0.95 & 1.36 \\
IQ3\_XXS & 2514 & 2872 & 804 & 653 & 12380 & 15722 & 1.15 & 1.54 \\
IQ3\_S & 2799 & 3000 & 757 & 676 & 14839 & 16496 & 1.14 & 1.49 \\
Q3\_K\_XL & 2240 & 2701 & 679 & 658 & 11211 & 14071 & 0.66 & 0.97 \\
\midrule
\multicolumn{9}{c}{\textbf{MMMU-Pro}} \\
\multirow{2}{*}{Decode format} & \multicolumn{2}{c|}{Mean tokens} & \multicolumn{2}{c|}{Median tokens} & \multicolumn{2}{c|}{95th percentile} & \multicolumn{2}{c}{Truncated (\%)} \\
 & WO & Pref. & WO & Pref. & WO & Pref. & WO & Pref. \\
\midrule
IQ1\_S & 14579 & 6582 & 10966 & 2393 & 32768 & 29953 & 24.57 & 4.16 \\
IQ1\_M & 8013 & 6875 & 3702 & 2459 & 32768 & 32768 & 6.24 & 6.47 \\
IQ2\_XXS & 5294 & 7377 & 3236 & 3337 & 16669 & 32768 & 0.52 & 5.55 \\
IQ2\_S & 4695 & 7118 & 2456 & 2930 & 15780 & 30445 & 0.40 & 4.34 \\
Q2\_K\_XL & 5094 & 6079 & 2823 & 2688 & 16736 & 25067 & 0.58 & 2.77 \\
IQ3\_XXS & 5400 & 6850 & 2844 & 2852 & 19363 & 29873 & 1.27 & 4.10 \\
IQ3\_S & 5860 & 6911 & 3212 & 3172 & 20578 & 28562 & 1.45 & 3.82 \\
Q3\_K\_XL & 5389 & 7156 & 2676 & 3305 & 19728 & 29416 & 1.39 & 3.53 \\
\bottomrule
\end{tabular}

%% file: tables/attn-backend.tex
\begin{tabular}{lrrl}
\toprule
Backend & $S{=}8192$ & $S{=}32768$ & Used for \\
\midrule
SDPA, \texttt{is\_causal} & \textbf{3.14} & \textbf{50.6} & global layers \\
FA2 varlen, full causal & 3.75 & 53.0 & - \\
FA4 (\texttt{flash\_attn.cute}) & 3.06 & 49.8 & - \\
FlexAttention, causal mask & - & 123.2 & - \\
\midrule
FlexAttention, sliding block mask & 2.03 & 8.7 & - \\
FA4, native \texttt{window\_size} & 3.02 & 49.0 & - \\
FA2 varlen, native window & \textbf{1.47} & \textbf{6.0} & local layers \\
\bottomrule
\end{tabular}

%% file: tables/prefill-per-model.tex
\begin{tabular}{lrrrrrr}
\toprule
\multirow{2}{*}{Model} & \multicolumn{3}{c}{$S{=}16384$} & \multicolumn{3}{c}{$S{=}32768$} \\
\cmidrule(lr){2-4} \cmidrule(lr){5-7}
 & BF16 (ms) & NVFP4 & +ODP & BF16 (ms) & NVFP4 & +ODP \\
\midrule
Qwen3-0.6B & 634 & 1.13$\times$ & 1.12$\times$ & 1984 & 1.08$\times$ & 1.05$\times$ \\
Qwen3-1.7B & 1044 & 1.24$\times$ & 1.19$\times$ & 2887 & 1.15$\times$ & 1.15$\times$ \\
Qwen3-4B & 2560 & 1.37$\times$ & 1.34$\times$ & 7335 & 1.14$\times$ & 1.15$\times$ \\
Qwen3-8B & 3801 & 1.49$\times$ & 1.47$\times$ & 9563 & 1.24$\times$ & 1.24$\times$ \\
\midrule
Gemma-3-270M & 124 & 1.22$\times$ & 1.16$\times$ & 281 & 1.14$\times$ & 1.11$\times$ \\
Gemma-3-1B & 473 & 1.45$\times$ & 1.34$\times$ & 999 & 1.43$\times$ & 1.41$\times$ \\
Gemma-3-4B & 1548 & 1.59$\times$ & 1.49$\times$ & 3231 & 1.53$\times$ & 1.49$\times$ \\
Gemma-3-12B & 4650 & 1.67$\times$ & 1.58$\times$ & 9607 & 1.59$\times$ & 1.55$\times$ \\
\bottomrule
\end{tabular}

%% file: tables/nvfp4-breakdown.tex
\begin{tabular}{lrrrr}
\toprule
Component & BF16 (ms) & NVFP4 (ms) & Speedup & \% NVFP4 \\
\midrule
\multicolumn{5}{l}{Gemma-3-12B: 48 layers, 40 sliding attention / 8 global attention} \\
\midrule
qkv & 10.23 & 5.57 & 1.84$\times$ & 10 \\
o & 6.51 & 1.46 & 4.47$\times$ & 3 \\
gate\_up & 38.47 & 11.66 & 3.30$\times$ & 21 \\
down & 20.26 & 6.32 & 3.21$\times$ & 11 \\
MLP non-linearity & 6.00 & 5.19 & 1.16$\times$ & 9 \\
Activation quant.\ (standalone) & - & 2.24 & - & 4 \\
Norms + RoPE + residual & 6.48 & 16.36 & 0.40$\times$ & 29 \\
Attention & 7.40 & 7.38 & 1.00$\times$ & 13 \\
\midrule
Device busy (sum of kernels) & 95.35 & 56.17 & 1.70$\times$ & 100 \\
Launch/idle gap & -0.24 & -0.24 & - & - \\
Layer (wall clock) & 95.12 & 55.94 & 1.70$\times$ & - \\
\textbf{Full stack, per layer} & \textbf{96.87} & \textbf{58.09} & \textbf{1.67}$\times$ & - \\
\midrule
\multicolumn{5}{l}{Qwen-3-8B: 36 layers, all global attention} \\
\midrule
qkv & 8.16 & 3.68 & 2.21$\times$ & 6 \\
o & 6.27 & 1.59 & 3.95$\times$ & 2 \\
gate\_up & 33.68 & 9.72 & 3.46$\times$ & 15 \\
down & 18.54 & 4.63 & 4.01$\times$ & 7 \\
MLP non-linearity & 4.79 & 4.19 & 1.14$\times$ & 6 \\
Activation quant.\ (standalone) & - & 2.34 & - & 4 \\
Norms + RoPE + residual & 6.44 & 15.81 & 0.41$\times$ & 24 \\
Attention & 25.59 & 24.59 & 1.04$\times$ & 37 \\
\midrule
Device busy (sum of kernels) & 103.46 & 66.54 & 1.55$\times$ & 100 \\
Launch/idle gap & -0.08 & -0.01 & - & - \\
Layer (wall clock) & 103.38 & 66.54 & 1.55$\times$ & - \\
\textbf{Full stack, per layer} & \textbf{105.57} & \textbf{71.09} & \textbf{1.49}$\times$ & - \\
\bottomrule
\end{tabular}